\documentclass[a4paper,fleqn]{cas-dc}

\usepackage[numbers]{natbib}
\usepackage{natbib}
\usepackage[utf8]{inputenc} 
\usepackage[T1]{fontenc}    
\usepackage{hyperref}       
\usepackage{url}            
\usepackage{booktabs}       
\usepackage{graphicx}       
\usepackage{amsfonts}       
\usepackage{amsmath, amssymb}
\usepackage{nicefrac}       
\usepackage{microtype}      
\usepackage{xcolor}         
\usepackage{multirow}
\usepackage{xcolor}
\usepackage{longtable}
\usepackage{array}
\usepackage{makeidx}
\usepackage{overpic} 
\usepackage{rotating} 
\usepackage{booktabs} 
\usepackage{adjustbox} 
\usepackage[dvipsnames]{xcolor}
\definecolor{revisionteal}{RGB}{0,105,140}
\newcommand{\rev}[1]{{\color{black}#1}}
\begin{document}
\let\WriteBookmarks\relax
\def\floatpagepagefraction{1}
\def\textpagefraction{.001}
\shorttitle{HiBerNAC: Hierarchical \rev{Brain-inspired} Robotic Neural Agent Collective for Disentangling Complex Manipulation}
\shortauthors{H. Zhang et~al.}

\title {HiBerNAC: Hierarchical \rev{Brain-inspired} Robotic Neural Agent Collective for Disentangling Complex Manipulation}                      



\author[1]{Heng Zhang}
\cormark[1]
\fnmark[1]
\ead{heng.zhang@iit.it}
    \affiliation[1]{Human-Robot Interfaces and Interaction Lab, Istituto Italiano di Tecnologia, Genoa, 16163, Italy.}

\author[2]{Hongjun Wu}
\fnmark[2]
\ead{hwu121@jh.edu}


\affiliation[2]{Department of Computer Science, Johns Hopkins University, Baltimore, MD 21218, USA}

\author[3]{Pengsong Zhang}
\fnmark[3]
\ead{pengsong.zhang@mail.utoronto.ca}
\affiliation[3]{Department of Mechanical and Industrial Engineering, University of Toronto, Toronto, M5S 1A1, Canada}

\author[1]{Jin Wang}
\fnmark[4]
\ead{Wang.Jin@iit.it}
\affiliation[4]{Humanoids and Human-Centered Mechatronics (HHCM), Istituto Italiano di Tecnologia, Genoa, 16163, Italy.}

\author[5]{Cong Wang}
\fnmark[5]
\ead{Wang.Jin@iit.it}
\affiliation[5]{Massachusetts General Hospital, Harvard Medical School, Boston, MA 02114, USA}

\cortext[cor1]{Corresponding author}

\begin{abstract}
Recent advances in multimodal vision language action (VLA) models have revolutionized traditional robot learning, enabling systems to interpret vision, language, and action in unified frameworks for complex task planning. 
However, mastering complex manipulation tasks remains an open challenge, constrained by limitations in persistent contextual memory, multi-agent coordination under uncertainty, and dynamic long-horizon planning across variable sequences.
\rev{To address this challenge, we draw conceptual inspiration from principles of neural organization and hierarchical decision-making, using them as architectural design metaphors rather than as claims of biologically faithful neural emulation.} 
We present \textbf{HiBerNAC}, a \rev{hierarchical brain-inspired robotic neural agent collective} that combines: (1) multimodal VLA planning and reasoning with (2) \rev{brain-inspired} reflection and multi-agent mechanisms, \rev{organized as a modular multi-agent architecture conceptually motivated by principles of neural organization,} specifically designed for complex robotic manipulation tasks.
By leveraging \rev{brain-inspired} functional modules with decentralized multi-agent collaboration, our approach enables robust and enhanced real-time execution of complex manipulation tasks. In addition, the agentic system has scalable collective intelligence via dynamic agent specialization, adapting its coordination strategy to variable task horizons and complexity.
Through extensive experiments \rev{(conducted in a MuJoCo-based simulation benchmark)} on complex manipulation tasks compared with the state-of-the-art VLA models, we demonstrate that {HiBerNAC} reduces average long-horizon task completion time by 23\%, and achieves non-zero success rates \rev{(8--48\%)} on multi-path tasks where prior state-of-the-art VLA models consistently failed \rev{(differences validated by two-sided Welch's $t$-tests, $p<0.05$ for correction, OOD, multimodal, and long-horizon categories)}. \rev{These results suggest that principles of neural organization can serve as useful inductive biases for designing robotic manipulation systems.} More details are available at our  \href{https://hibernac.github.io}{\texttt{project website}}.
\end{abstract}
\begin{figure*}
    \centering
    \includegraphics[width=1\linewidth]{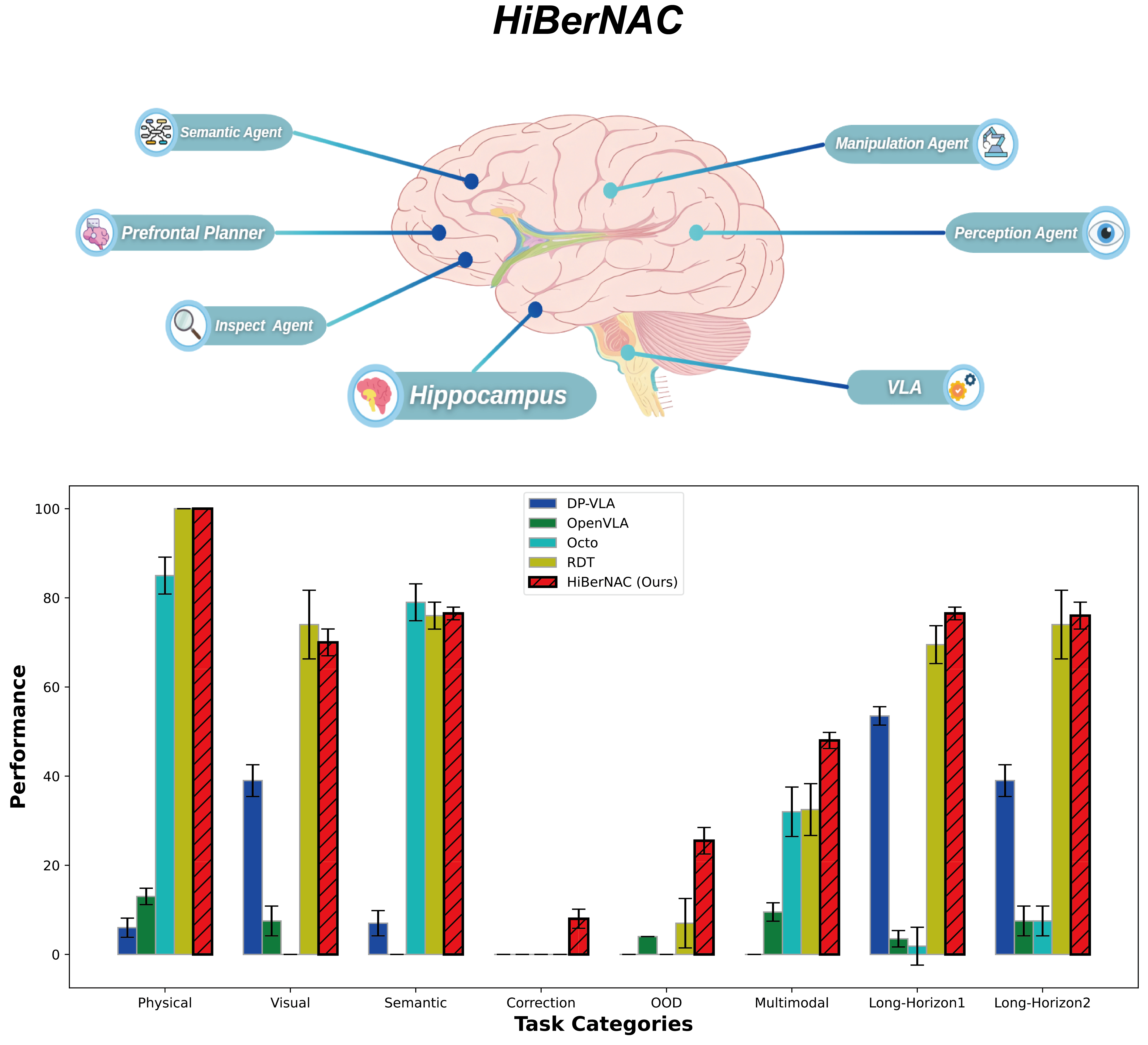}
    \caption{Evaluation of Performances in comparison with SOTA VLA models}
    \label{fig:Eva-performances}
\end{figure*}


\begin{highlights}
\item \rev{A hierarchical multi-agent architecture conceptually inspired by principles of neural organization.}
\item \rev{Multimodal VLA planning and reasoning with memory-guided reflection and decentralized multi-agent coordination.}
\item Designed for complex robotic manipulation tasks.
\end{highlights}

\begin{keywords}
Manipulation, Embodied AI, Agent, Vision-Language-Action model
\end{keywords}

\maketitle
\section{Introduction}
\label{sec:intro}
As robots take on increasingly complex tasks in daily life scenarios, building an adaptive and scalable system is a significant challenge \cite{mukherjee2022survey, hamann2021scalability}. While natural language instructions and textual environment descriptions offer an intuitive interface for robot interaction, they often require carefully engineered inputs and struggle with robustness in long-horizon tasks or in environments that are only partially known \citep{zeng2023large}. Multi-agent frameworks offer a more socially inspired approach to problem solving, shifting much of the cognitive load from individual reasoning to inter-agent communication~\citep{gronauer2022multi}. However, when relying on natural language as the medium, this communication becomes costly, requiring complex mechanisms for error correction and semantic verification to ensure alignment and correctness, which significantly compromises overall efficiency~\citep{mostajabdaveh2024optimization, li2024survey}. Recent advances in multimodal large language models have demonstrated unprecedented capabilities in understanding and generating cross-modal content~\citep{zhang2024mm}. Meanwhile, neuroscience-inspired approaches to artificial intelligence have shown promising results in creating more robust and adaptable systems~\citep{ liu2024cognitive}. The integration of these approaches with multi-agent systems presents a unique opportunity to advance robot learning. Traditional approaches to robot learning often lack the flexibility and adaptability seen in human cognition. While recent work in brain-inspired computing~\citep{lechner2020neural, zhang2020system} has shown promise in addressing these limitations, the integration of these approaches with modern AI systems remains challenging. Our work addresses this gap by combining neuroscience principles with state-of-the-art multimodal language models~\citep{palm2} in a coordinated multi-agent framework.

Despite these favorable advances, the development of intelligent robots capable of learning and executing complex skills, achieving human-like proficiency in long-horizon tasks remains a fundamental challenge in robotics\citep{liu2023imitation, dalal2024plan}. Several key technical issues persist that hinder effective robot learning systems. Multimodal perception-action integration presents significant challenges, as combining diverse sensory inputs (visual, tactile, proprioceptive, audio) with language understanding necessitates sophisticated fusion architectures. Current approaches struggle with persistent contextual memory across variable sequences, dynamic planning for complex manipulation tasks, and real-time adaptation to unexpected situations~\citep{zhang2025safe}. Single-agent architectures lack the specialization necessary for complex cognitive functions, while pure language-based multi-agent systems generate excessive communication overhead\citep{tran2025multi}. Furthermore, existing systems typically operate at uniform processing frequencies, unable to balance the computational efficiency of reactive control with the contextual depth of deliberative planning. Multi-agent coordination requires robust communication protocols and conflict resolution strategies to synchronize specialized agents for perception, planning, execution, and safety monitoring. Translating principles from brain function into computational models involves complex trade-offs between biological plausibility, interpretability, and efficiency. Additionally, practical deployment demands real-time control with fast processing cycles, skill transfer and generalization to novel contexts without extensive retraining, and ensure safety in physical environments. These challenges are further compounded in long-horizon tasks or partially observable environments, where traditional approaches often fail to maintain coherent behavior or adapt to unexpected situations. The lack of integrated solutions addressing these interdependent issues has limited the deployment of truly adaptive robots in complex real-world scenarios.

To address these challenges, we propose \textbf{HiBerNAC}, a novel framework that integrates multimodal VLA planning with neuro-inspired multi-agent collaboration for complex manipulation tasks, see fig.~\ref{fig:Eva-performances}. 
The cognitive processes of humans rely on the distributed and modular structure of the brain, where different functional areas (such as the sensory cortex, motor cortex, hippocampus, etc.) work together through neurons to accomplish complex tasks of perception, memory, and action.
Our brain-inspired architecture models robotic cognition as a multimodal multiagent system \rev{whose modular organization is conceptually motivated by neural specialization in the human brain (the biological labels serve as design metaphors rather than mechanistic claims)}. For complex long-horizon tasks, the system employs a Multi-agent Neural Structure operating at $1 \times 10^{-2}\ \mathrm{Hz}$ 
featuring a Prefrontal Planner for high-level cognition and a Hippocampus module for episodic memory. This deliberative pathway communicates with an Asynchronous Pipeline implementing a DBHTN(DAG) planner with complementary memory systems for hierarchical task management. Simple reactive tasks leverage a streamlined pathway through the Hippocampus at $1 \times 10^{-1}\ \mathrm{Hz}$ or utilize the Reactive VLA system operating at 1e2Hz for low-level control based on open-source pretrained models. This multi-pathway, multi-frequency architecture enables both long-horizon reasoning and immediate environmental response, dynamically allocating computational resources based on task complexity and familiarity. Through this integrated approach, \textbf{HiBerNAC} achieves robust execution of complex manipulation tasks while demonstrating scalable collective intelligence via dynamic agent specialization.

\begin{figure*}
    \centering
    \includegraphics[width=1\linewidth]{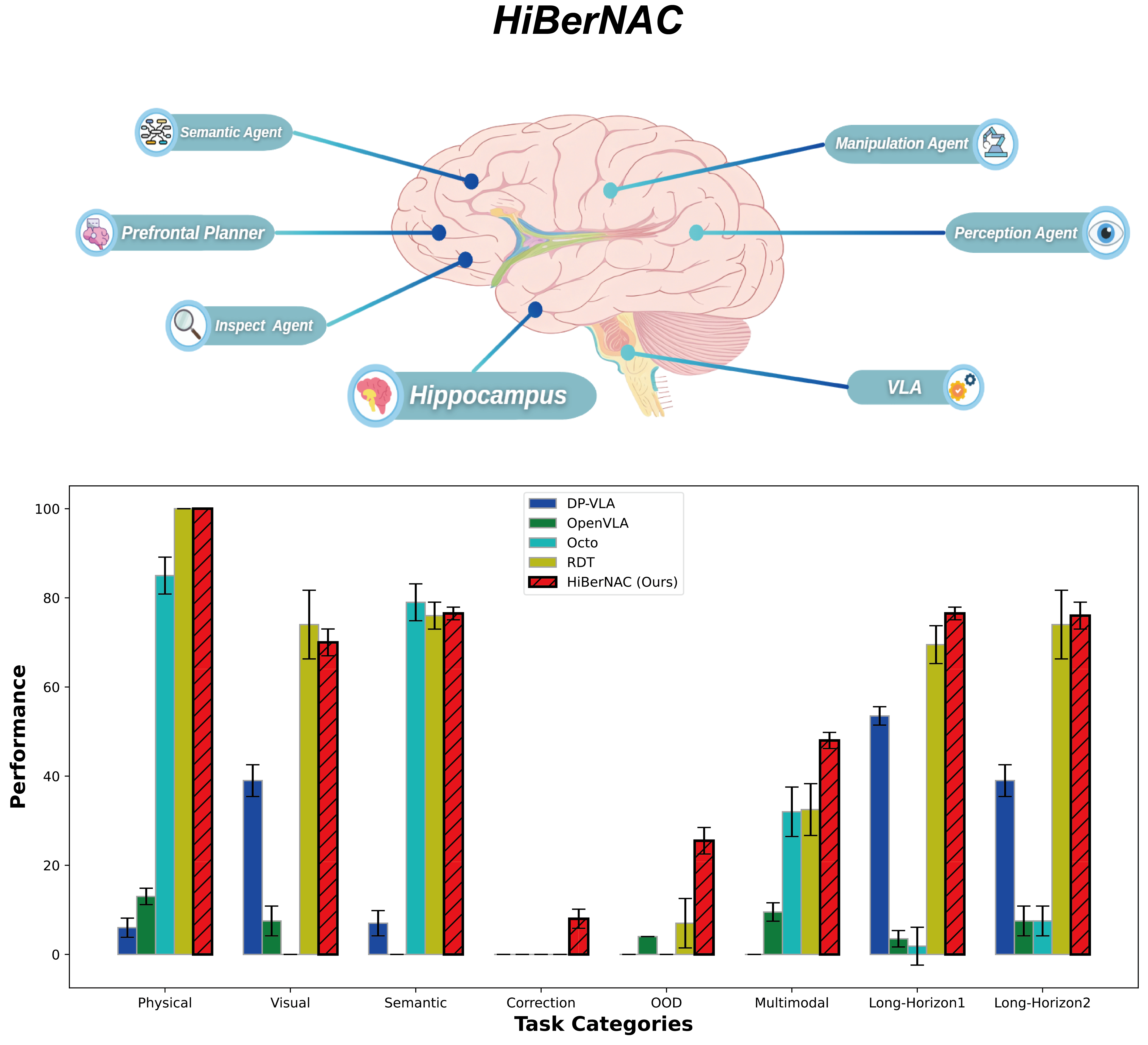}
    \caption{Evaluation of Performances in comparison with SOTA VLA models}
    \label{fig:Eva-performances}
\end{figure*}

Through extensive experiments on the 7-DoF Franka robot platform \rev{conducted entirely in MuJoCo-based simulation.}, we demonstrate that \textbf{HiBerNAC} achieves significant improvements in skill acquisition, generalization, and adaptation. Our framework achieves a 23\% average reduction in execution time for long-horizon tasks, and a \rev{16.5 percentage-point absolute improvement (48.5\% vs.\ 32\% for the strongest baseline RDT)} in success rate for multimodal task completion, outperforming existing state-of-the-art approaches.

\section{Related Work}
\label{sec:related}
\textbf{Traditional Robot-Skill Learning}.
Dynamic Movement Primitives: Learnable motion primitives that generalize across variations in object properties and environmental conditions~\citep{liao2022dynamic,lu2024dynamic, jiang2025cdpmm}.
Interactive Learning: Integrating human feedback and corrections during skill acquisition to accelerate training~\citep{chetouani2021interactive, cansev2021interactive}.
Multi-Task Skill Transfer: Leveraging shared representations across tasks to enable rapid adaptation to new scenarios~\citep{upadhyay2024sharing}.
Few-shot Imitation: Learning complex motions from minimal human demonstrations using pretrained motion priors~\citep{ren2025motion, vecerik2024robotap}.
Hierarchical Learning: Combining high-level task planning with low-level motor skills~\citep{gehring2021hierarchical, liu2023hierarchical, rao2021learning}.
Adaptive Control: Online adaptation to environmental changes and perturbations~\citep{kazim2021disturbance, fan2021dynamic, ebrahimi2021adaptive}.
RL-based Skill Policy: RL-based policy learning has been widely adopted for robot skill learning~\citep{hansen2022visuotactile,sun2022fully,10517611}, enabling autonomous agents to acquire complex behaviors through trial-and-error interactions with their environment~\citep{luo2023action,zhang2025bresa}.

\textbf{Multimodal Large Language Models for Manipulation}. Recent studies have demonstrated the interpretability, flexibility, scalability, and effectiveness of leveraging LLMs in performing autonomous planning and execution in embodied robotic environments~\cite{wang2024survey, xi2025rise, sumers2023cognitive, song2023llm, huang2022language, huang2022inner}. Grounding natural language in action plans is an active area of research within embodied agent planning~\cite{brohan2023can, huang2023grounded, rana2023sayplan, liu2023grounding, 10802344, wang2024hypermotion}. Research has also explored direct fine-tuning of language models to generate action output~\cite{suglia2021embodied, pashevich2021episodic, sharma2021skill}. Multi-agent systems for coordination planning and tasks decomposition~\citep{kannan2024smart,singh2024twostep}. More recent works focus on fully decentralized multi-agent systems and research the potential of cooperation among agents using LLMs~\cite{mandi2024roco, chen2024scalable, zhang2023building, wang2024probabilistically, nayak2024long}. However, multi-agent LLM systems' challenges persist, including ensuring coherent conversation and state tracking among agents, minimizing communication overhead while maintaining robust planning strategies for complex manipulation, and balancing centralized knowledge with distributed execution.

\textbf{Vision-Language-Action models for generalized manipulation} have achieved remarkable progress in bridging visual and semantic understanding, especially with the help of the dataset of Open X-Embodiment~\cite{o2024open} and transformer-based skill learning structure~\citep{firoozi2023foundation}. Advanced works such as 
OpenVLA~\citep{kim2024openvla}, which demonstrated that pretrained vision-language models can achieve better performance in robotic manipulation through grounding language instructions in visual action plans. OpenVLA-oft~\citep{kim2025fine} proposed an Optimized Fine-Tuning (OFT) recipe integrating parallel decoding, action chunking, a continuous action representation, and a simple $L1$ regression-based learning objective to improve inference efficiency altogether. DP-VLA~\citep{han2024dual} achieves faster inference and higher task success rates due to the proposed dual-process system. Octo~\citep{team2024octo} and RDT~\citep{liu2024rdt} enhanced generalization in task categorization and robot embodiments. More recent related works \citep{yue2024deer,gbagbe2024bi,fan2025interleave,zawalski2024robotic,zhang2026tacvla,deng2025graspvla,zhang2026compliantvla,bu2025univla} demonstrate the promises of VLA further from different perspectives. However, most of them remain challenges in long-horizon tasks, real-time performance, and efficient reasoning.

{Neuroscience-inspired Computing}. Recent advances in neuromorphic computing and brain-inspired algorithms have led to more efficient and adaptive AI systems. 
Key developments can be mainly divided into two groups:
     (1). Neuroscience-inspired Learning for embodied agent: New algorithms such as Forward Propagation Through Time (FPTT) enable online training of recurrent spiking neural networks (SNNs). Neural Brain~\citep{liu2025neural}integrates multimodal active sensing, perception-cognition-action function for embodied agents.
     (2). Spike-based Models: \cite{khacef2023spike,hong2024spaic} designed brain-inspired models and algorithms and SpikeGPT demonstrates the potential to merge language model capabilities with neuromorphic computing~\citep{zhu2023spikegpt}.
However, there is few work that designs a systemic mechanism inspired by a neuroscience perspective for complex robot manipulation.

\section{Methodology}
\label{sec:method}

\begin{figure*}[tbh]
    \centering
    \includegraphics[trim=0.5cm 0cm 0cm 0.5cm,width=1\linewidth]{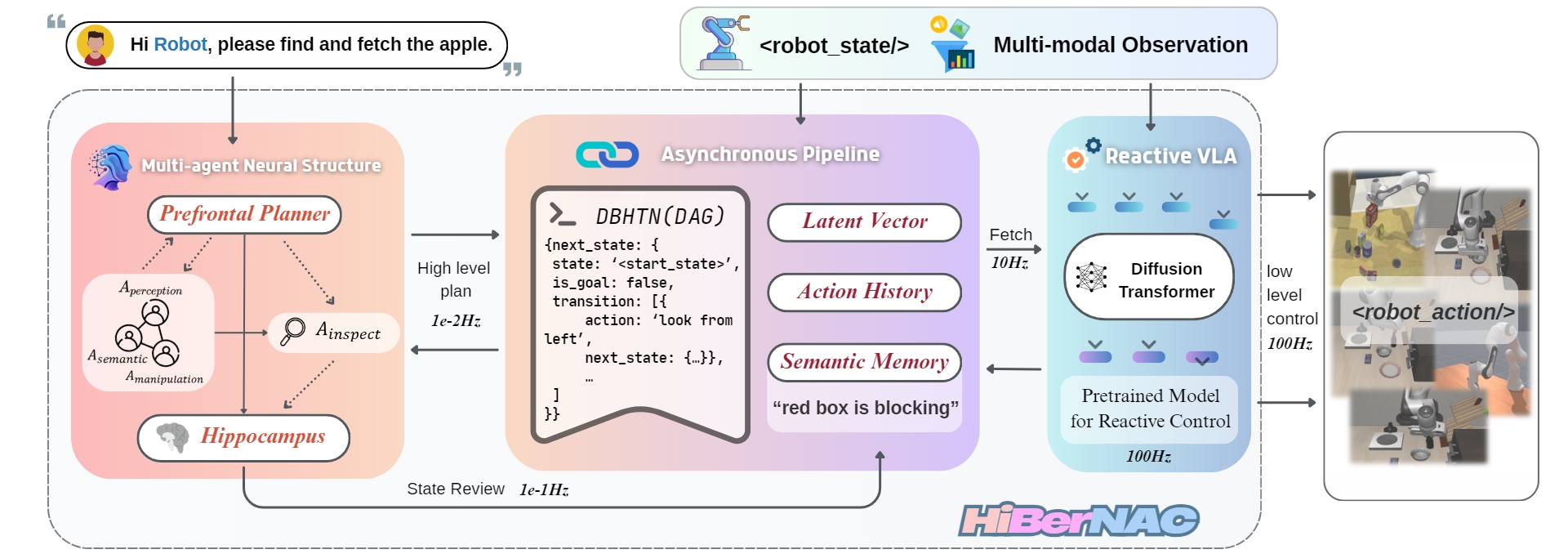}
    \caption{Overall framework of \textbf{HiBerNAC}. Three main core modules: Left. A multi-agent neural structure for high-level planning based on language input; Middle. An asynchronous hierarchical task pipeline for neural agent collective; and Right. A reactive VLA model for real-time robotic manipulation execution.}
    \label{fig:framework}
\end{figure*}

Our brain-inspired planning and manipulation framework models neural cognition structure as a multimodal multiagent system, \rev{whose modular organization is conceptually motivated by neural specialization in the human brain. We emphasize that the biological region labels (prefrontal cortex, hippocampus, inferior olive, cerebellum, etc.) are used as \emph{design metaphors} to motivate modular specialization and multi-frequency operation; they do not imply biophysical equivalence or constitute neuroscience validation. Concretely, three architectural decisions were directly motivated by these analogies: (i) the Hippocampus Module's gated memory decay (Eq.~\ref{eq:hm}) was inspired by hippocampal consolidation; (ii) the Inspection Agent's error-feedback loop was motivated by the error-correction role of the inferior olive--cerebellar axis; and (iii) the asynchronous multi-frequency operation ($10^{-2}$~Hz / $10^{-1}$~Hz / $10^{2}$~Hz) reflects the brain's distinct timescales for deliberative versus reflexive processing,} as illustrated in Figure~\ref{fig:framework}. For complex and long-horizon tasks, the system dynamically activates distinct agents for processing language, vision, and episodic memory inputs. These agents function analogously to sensory cortices, extracting salient features and contextual information. The fused multimodal representations are routed to a central planning module inspired by the prefrontal cortex, where high-level decision-making and task decomposition occur. This output is further refined through a correction mechanism modeled on the inferior olivary nucleus, introducing predictive error feedback for robust adaptation. In contrast, simple or reactive tasks bypass high-level planning and instead utilize a streamlined single-agent pathway akin to spinal or reflex arcs in biological systems. These rely on a 'past work memory' module that recalls previously successful execution patterns, slightly modified to fit current contexts, ensuring both rapid response and task relevance. Task routing between these two pathways is governed by a context-aware task classifier that estimates cognitive load and task complexity. This architecture supports real-time adaptability and efficiency, integrating deliberative and reflexive behaviors. The structure graph accompanying this section outlines these modules and their interactions, reflecting both the functional and anatomical parallels to the human brain. The system's implementation integrates three interacting components: (1) a multi-agent neural structure for high-level planning, (2) an asynchronous pipeline for hierarchical task management, and (3) a reactive VLA system for real-time control execution.

\subsection{Multi-agent Neural Structure}
\label{subsec:neural_structure}

\begin{figure*}[tbh]
    \centering
    \includegraphics[trim=4.5cm 0cm 4.5cm 0cm, width=1\linewidth]{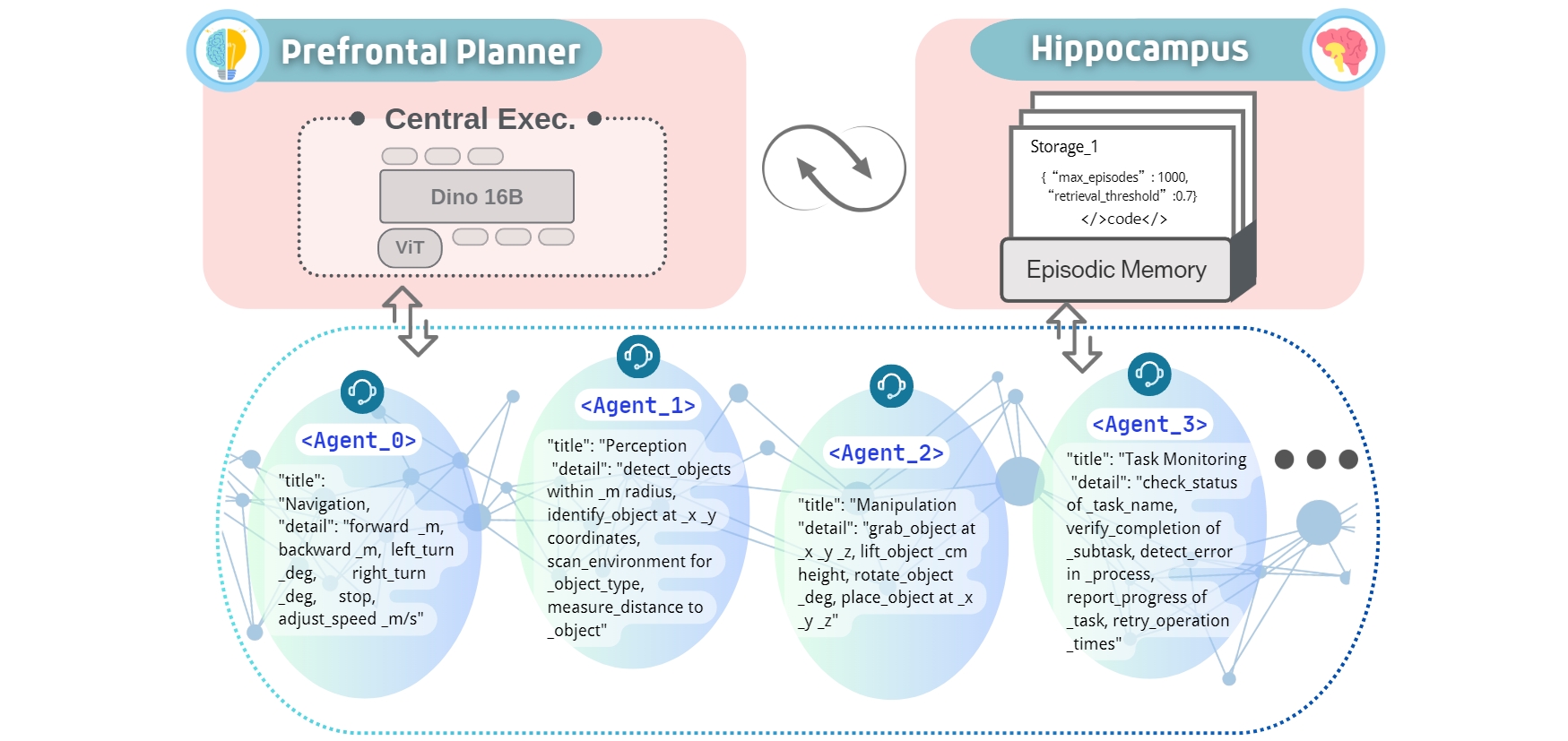}
    \caption{Multi-agent Neural Architecture: A collective of neural agents designed for scalable and adaptive operation across variable task horizons and complexity levels, achieved through dynamic agent specialization. }
    \label{fig:multi-agent}
\end{figure*}

The multi-agent neural structure is the deliberative core of \textbf{HiBerNAC}, operating at $1 \times 10^{-2}\ \mathrm{Hz}$ for high-level planning and contextual understanding, as shown in Fig.~\ref{fig:multi-agent}. Specifically, the Prefrontal Planner (\texttt{PFP}), analogous to the prefrontal cortex, performs high-level reasoning to decompose tasks (e.g., "find and fetch the apple") into structured subtasks. Surrounding the PFP, specialized agents process sensory and cognitive information: the Perception Agent (\texttt{PA}) extracts environmental features, the Semantic Agent (\texttt{SA}) interprets instructions, the Manipulation Agent (\texttt{A\_manipulation}) translates goals into actions, and the Inspection Agent (\texttt{IA}) monitors execution. The Hippocampus Module (\texttt{HM}) implements episodic memory, storing past task contexts at $1 \times 10^{-1}\ \mathrm{Hz}$, enabling more rapid decision-making than PFP for reactive tasks by bypassing planning.

{Synaptic Interaction Circuitry.} Inspired by functional connectivity in neural circuits, \textbf{HiBerNAC} models interactions between agents, each representing a brain region, using a general function that captures both individual processing and collective coordination:
\begin{equation}
\mathbf{o}_t^i = A_i(\mathbf{x}_t^i, \mathbf{s}_t^i, \mathbf{m}_t; \Theta_i) + \sum_{j \neq i} \mathbf{F}_{ij} \cdot \mathbf{o}_{t-1}^j
\label{eq:collaboration}
\end{equation}
Here, \(\mathbf{o}_t^i\) denotes the output of agent \(i\) at time \(t\), such as a task plan, sensory feature, or action, analogous to aggregated neural signals from a brain region. The term \(\mathbf{x}_t^i\) represents external inputs, like sensory data or task instructions, \(\mathbf{s}_t^i\) is the agent’s internal state, and \(\mathbf{m}_t \in \mathbb{R}^{d_m}\) is a shared memory state, akin to the hippocampus’s episodic memory. The parameters \(\Theta_i\) are pretrained, adapted from VLMs or LLMs, encapsulating internal neural connections within each agent. The interaction term \(\sum_{j \neq i} \mathbf{F}_{ij} \cdot \mathbf{o}_{t-1}^j\), where \(\mathbf{F}_{ij}\) is a functional connectivity matrix, models signal propagation between agents, \rev{drawing a structural analogy to region-to-region pathways in the brain rather than claiming biological equivalence}. This framework enables \textbf{HiBerNAC} to adaptively coordinate agents, much like the nervous system integrates distributed computations.

{Cortical Processing Units.} \textbf{HiBerNAC}’s agents, inspired by the brain’s neural specialization, \rev{form modular units conceptually motivated by cortical hierarchy and subcortical feedback}, collaborating to execute tasks like a robot fetching an apple. Integrating multimodal sensory inputs, semantic context, and episodic memory, these agents dynamically route tasks between deliberative planning and reflexive actions, akin to the brain’s switch between cortical control and spinal reflex arcs, ensuring adaptability and efficiency.

{Prefrontal Planner (\texttt{PFP}).} Akin to the prefrontal cortex, serves as the system’s strategic command center, decomposing complex instructions into structured subtasks. For instance, given the command ``find and fetch the apple,'' the \texttt{PFP} generates a sequence of subtasks, such as ``scan environment,'' ``locate apple,'' and ``grasp object.'' Its operation is formalized as:
\begin{equation}
\mathbf{p}_t = A_{\text{PFP}}(\mathbf{s}_t, \mathbf{m}_t, \mathbf{h}_{t-1}^{\text{SA}}; \psi)
\label{eq:pfp}
\end{equation}
In this equation, \(\mathbf{p}_t \in \mathbb{R}^{d_p}\) is the task plan, \(\mathbf{s}_t \in \mathbb{R}^{d_s}\) is the current environmental state, \(\mathbf{m}_t\) is the shared memory, \(\mathbf{h}_{t-1}^{\text{SA}} \in \mathbb{R}^{d_h}\) is the prior semantic context from the Semantic Agent, and \(\psi\) denotes pretrained parameters. The \texttt{PFP} integrates these inputs, much like the prefrontal cortex coordinates sensory and memory signals for decision-making, and its output \(\mathbf{p}_t\) guides the Manipulation Agent via \(\mathbf{W}_{\text{PFP,MA}}\), while receiving corrective feedback from the Inspection Agent through \(\mathbf{W}_{\text{IA,PFP}}\).

{Perception Agent (\texttt{PA}).} Modeled after sensory cortices, processes multimodal inputs, such as visual and auditory data, to extract environmental features, similar to how the brain interprets sights and sounds. Using a pretrained VLM, the \texttt{PA} fuses these inputs into a coherent representation:
\begin{equation}
\mathbf{z}_t = A_{\text{PA}}\left( \sum_{m \in \mathcal{M}} \mathbf{W}_m \cdot \text{Emb}_m(\mathbf{o}_t^m; \theta_m); \theta \right)
\label{eq:pa}
\end{equation}
Here, \(\mathbf{z}_t \in \mathbb{R}^{d_z}\) is the feature vector, \(\mathbf{o}_t^m\) represents modality-specific observations (e.g., camera images, task text), \(\text{Emb}_m(\cdot; \theta_m)\) embeds observations with pretrained parameters \(\theta_m\), \(\mathbf{W}_m\) are fusion weights, and \(\theta\) is the VLM’s pretrained parameters. The \texttt{PA}’s output \(\mathbf{z}_t\) flows to the Semantic Agent via \(\mathbf{W}_{\text{PA,SA}}\), enabling contextual interpretation, much like sensory cortices relay signals to association areas.

{Semantic Agent (\texttt{SA}).} Analogous to language and association cortices, interprets task instructions by applying attention mechanisms to the \texttt{PA}’s features \(\mathbf{z}_t\), shared memory \(\mathbf{m}_t\), and its prior state \(\mathbf{h}_{t-1}^{\text{SA}}\). It generates a semantic state \(\mathbf{h}_t^{\text{SA}}\), which supports the \texttt{PFP}’s planning and the Inspection Agent’s monitoring, akin to the brain’s ability to comprehend language and context. Leveraging a pretrained VLM with multi-head attention, the \texttt{SA} ensures robust instruction understanding, broadcasting its output through \(\mathbf{W}_{\text{SA,PFP}}\) and \(\mathbf{W}_{\text{SA,IA}}\).

{Manipulation Agent (\texttt{MA}).} Inspired by the motor cortex, translates the \texttt{PFP}’s plan \(\mathbf{p}_t\) into executable actions \(\mathbf{a}_t\), such as moving a robotic arm to grasp an apple. Operating on the current state \(\mathbf{s}_t\) and memory \(\mathbf{m}_t\), the \texttt{MA} employs a pretrained model with contextual attention to maintain action continuity, receiving direct commands from the \texttt{PFP} via \(\mathbf{W}_{\text{PFP,MA}}\) and adjustments from the Inspection Agent’s feedback, similar to how the motor cortex executes movements refined by cerebellar input.

{Inspect Agent (\texttt{IA}).} Modeled after the cerebellum and inferior olivary nucleus, acts as an error calibrator, monitoring task execution by comparing the \texttt{PFP}’s plan \(\mathbf{p}_t\) with the \texttt{SA}’s semantic state \(\mathbf{h}_t^{\text{SA}}\). It produces a monitoring vector \(\mathbf{c}_t\), which feeds back to the \texttt{PFP} through \(\mathbf{W}_{\text{IA,PFP}}\), enabling adaptive replanning, much like the cerebellum corrects motor errors to ensure smooth coordination.

{Hippocampus Module (\texttt{HM}).} \rev{Conceptually inspired by hippocampal memory function, the module} serves as the system’s memory archive, storing and retrieving task contexts at $1 \times 10^{-1}\ \mathrm{Hz}$ to support rapid decision-making, particularly in reactive tasks. Its dynamics are defined as:
\begin{equation}
\mathbf{m}_t = -\alpha_m \mathbf{m}_{t-1} + A_{\text{HM}}(\mathbf{s}_t, \mathbf{z}_t, \mathbf{m}_{t-1}; \gamma)
\label{eq:hm}
\end{equation}
Here, \(\mathbf{m}_t \in \mathbb{R}^{d_m}\) is the memory state, \(\alpha_m\) is the decay rate, \(\mathbf{s}_t\) and \(\mathbf{z}_t\) are state and feature inputs, \(\mathbf{m}_{t-1}\) is the prior memory, and \(\gamma\) denotes pretrained parameters. The \texttt{HM} broadcasts \(\mathbf{m}_t\) to all agents via \(\mathbf{W}_{\text{HM},i}\), mimicking hippocampal memory consolidation, enabling the system to recall past actions, much like the brain retrieves muscle memory for riding a bicycle.

\subsection{Asynchronous Pipeline}
The asynchronous pipeline bridges \textbf{HiBerNAC}’s deliberative multi-agent structure and reactive control system, managing information flow in a neural-like relay. The Decomposition-Based Hierarchical Task Network (DBHTN) planner, visualized as a directed acyclic graph (DAG) in Fig.~\ref{fig:framework}, decomposes \texttt{PFP}’s high-level plans into structured action sequences, such as ``find and fetch the apple'' into states (``start\_state'') and transitions (``look from left''). Memory systems include a latent vector \(\mathbf{l}_t \in \mathbb{R}^{d_l}\) for environmental states, action history \(\mathbf{h}_a = [\mathbf{a}_{t-1}, \mathbf{a}_{t-2}, \dots, \mathbf{a}_{t-k}]\), and semantic memory \(\mathbf{m}_s \in \mathbb{R}^{d_s}\) for task concepts, enabling asynchronous updates. A state review mechanism at $1 \times 10^{-1}\ \mathrm{Hz}$ adjusts plans by comparing states. The update is:
\begin{equation}
\mathbf{l}_{t+1} = A_{\text{pipeline}}(\mathbf{l}_t, \mathbf{a}_t, \mathbf{m}_{\text{task}}, \mathbf{T}_{\text{DAG}}(\mathbf{p}_t); \theta_{\text{pipe}}) + \lambda \cdot \text{DBN}(\mathbf{e}_t, \mathbf{o}_t; \phi)
\label{eq:pipeline}
\end{equation}
A dynamic Bayesian network (DBN), modeling states and observations with temporal dependencies, infers hidden states for multi-step planning, with an LLM reward function enhancing contextual inference via EM optimization. Fig.~\ref{fig:framework} illustrates the DAG and asynchronous flow.

\subsection{Reactive VLA (RVLA)}
\label{subsec:reactive_vla}

The RVLA handles low-level, real-time reactive actions, simulating the spinal cord or lower motor areas with high responsiveness. It operates at 100Hz, enabling immediate environmental responses based on robot state and multimodal observations, using an open-source pretrained VLA model to map sensory inputs to action outputs. It receives visual data, proprioceptive feedback, and state information from robot sensors, ensuring responsive execution while higher cognitive processes run at lower frequencies. The RVLA maintains bidirectional data exchange with the asynchronous pipeline, receiving "fetch" commands at $10Hz$ and sending execution feedback to enable adaptive replanning, forming a closed-loop system for integrating deliberative planning with reactive control.
\rev{
\begin{equation}
\begin{aligned}
\mathbf{u}_t &= A_{\text{RVLA}}(\mathbf{s}_t, \mathbf{a}_t, \mathbf{l}_t; \nu) \\
&\quad + \zeta \cdot \left( \text{Ctrl}(\mathbf{e}_t; \nu_{\text{Controller}})
      + \sigma \cdot \text{Var}_{\text{react}}(\mathbf{s}_t, \mathbf{l}_t) \right)
\end{aligned}
\end{equation}}
Here, \(\mathbf{u}_t \in \mathbb{R}^{d_u}\) is the reactive control output, \(\mathbf{s}_t\) is the current state, \(\mathbf{a}_t\) is the action from MA, \(\mathbf{l}_t\) is the latent vector from the pipeline, \(\nu\) and \(\nu_{\text{Controller}}\) are parameters, Ctrl indicates controller, \(\mathbf{e}_t = \mathbf{s}_t - \hat{\mathbf{s}}_t\) is the prediction error, \(\zeta\) and \(\sigma\) are coefficients controlling controllers adjustment and variance normalization, and \(\text{Var}_{\text{react}}\) ensures stability in dynamic environments. The RVLA operates at a fast timescale (\(\tau_{\text{react}} \approx 10 \, \text{ms}\)) and can leverage real-time VLA models.

\section{Experimental Implementation and Evaluation}
\label{sec:exp_eval}

In this section, we evaluate \textbf{HiBerNAC} across a spectrum of manipulation tasks, ranging from basic object interactions to semantically complex, long-horizon scenarios. Our experiments systematically assess: (1) its execution performance compared to state-of-the-art methods, (2) the benefits of dynamic multi-agent coordination for real-time error correction, and (3) its scalability in handling extended task sequences.

\subsection{Implementation Details}

{Experimental Setup}. We validate \textbf{HiBerNAC} on a 7-Dof Franka robot designed for complex manipulation tasks, equipped with dual-arm manipulation (7 DOF per arm), multi-finger hands with embedded tactile arrays, RGB-D cameras for stereo vision, force-torque sensors at key joints with an IMU for dynamic balance, a 16-channel microphone array, and on-board GPU accelerators for real-time processing, supported by a sensor suite including high-resolution cameras (2x 4K, 30Hz), tactile sensors (500Hz sampling), joint encoders (1kHz sampling), and an IMU (100Hz sampling) for multimodal capabilities in vision, touch, motion tracking, and balance. The software stack leverages a distributed computing framework with GPU-accelerated neural networks for object detection, semantic segmentation, pose estimation, command parsing, context summarization, and closed-loop motion control, using 100Hz real-time control loops, parallel pipelines for perception, planning, and action generation, and fault-tolerant execution with autorecovery triggers, where each agent operates as a separate process with independent state management, a message-passing interface, and CPU/GPU monitoring, enabling modular upgrades like adding a new Perception Agent without recompiling the stack. All experiments are conducted using a cluster equipped with two NVIDIA RTX 3090 GPUs (48GB VRAM), 32 vCPUs on an Intel Xeon Gold 6330 CPU (2.00GHz), 512GB RAM.
\textbf{HiBerNAC}’s skill library supports these tasks with modular primitives for locomotion (gait generation, navigation), object manipulation (grasping, in-hand manipulation with tactile feedback), tool use (tool recognition, force-sensitive control), task planning (goal decomposition, action sequencing, error detection, re-planning).

\textbf{Baselines}.
We conduct a comprehensive evaluation of our approach against leading VLA models, including DP-VLA~\citep{han2024dual}, OpenVLA~\citep{kim2024openvla}, Octo~\citep{team2024octo}, and RDT~\citep{liu2024rdt}, to rigorously assess its capabilities. Our experiments demonstrate consistent improvements in task success rates, surpassing state-of-the-art baselines by a notable margin (See Fig.~\ref{fig:Eva-performances} and Tab.~\ref{tab:metrics}).

\rev{\textbf{Baseline Evaluation Protocol.} All baseline models (DP-VLA, OpenVLA, Octo, RDT) are evaluated \emph{off-the-shelf} using their publicly released checkpoints, without task-specific fine-tuning. The purpose of this protocol is to assess the generalization capability of pretrained generalist VLA models under a common deployment setting, rather than to estimate their task-specific performance ceilings. We did not optimize hyperparameters beyond the defaults reported in the original papers. Accordingly, the reported results should not be interpreted as demonstrating that HiBerNAC necessarily outperforms optimally fine-tuned versions of these baselines. The low success rates observed on Correction and OOD tasks are consistent with the absence, in the evaluated baseline configurations, of explicit persistent memory, replanning, or dedicated error-correction mechanisms; however, task-specific fine-tuning, parameter-efficient adaptation, or benchmark-specific policy optimization may substantially improve baseline performance. A controlled comparison against such adapted baselines is therefore an important direction for future work in Section~\ref{sec:conclusion}.}

\textbf{Benchmark Tasks}. 
Specifically, we benchmark performance across eight diverse task categories, encompassing physical interaction, visual understanding, semantic reasoning, error correction, out-of-distribution (OOD) generalization, multi-modal integration, and two distinct long-horizon manipulation scenarios, highlighting the robustness and versatility of our method.

For more detailed implementation, baselines, and task settings, please refer to the Appendix.
\subsection{Main Results}

Our evaluation demonstrates that \textbf{HiBerNAC} outperforms state-of-the-art VLA models across diverse task categories, as summarized in Table~\ref{tab:metrics} and illustrated in Fig.~\ref{fig:Eva-performances}. In Physical tasks (Task 1), \textbf{HiBerNAC} achieves a 100\% success rate with zero standard deviation, matching RDT and surpassing DP (6\%), OpenVLA (13\%), and Octo (85\%), indicating exceptional robustness in basic physical interactions. For Visual tasks (Task 2), \textbf{HiBerNAC} attains a 70\% success rate (std: 3.02), slightly below RDT’s 74\% (std: 7.71), but significantly outperforms other baselines, suggesting potential for further enhancement in visual processing. In Semantic tasks (Task 3), \textbf{HiBerNAC}’s 76.5\% success rate (std: 1.41) is competitive with Octo (79\%) and RDT (76\%), with a lower standard deviation reflecting greater consistency.

\textbf{HiBerNAC} excels notably in specialized tasks. In Correction tasks (Task 4), it achieves an 8\% success rate (std: 2.14), far exceeding RDT (1.5\%) and others (0-0\%), highlighting its superior error correction capability. For OOD tasks (Task 5), \textbf{HiBerNAC}’s 25.5\% success rate (std: 2.98) outperforms RDT (7\%) and others (0-4\%), demonstrating robust generalization to unseen environments. In Multimodal tasks (Task 6), \textbf{HiBerNAC} reaches 48\% (std: 1.81), significantly ahead of RDT (32\%) and others (0-9.5\%), underscoring its strength in integrating diverse inputs. In {Long-Horizon} tasks, \textbf{HiBerNAC} achieves 76.5\% (Task 7) and 76\% (Task 8), surpassing RDT by 7\% and 2\%, respectively, affirming its scalability in complex, multi-step scenarios.

Beyond success rates, \textbf{HiBerNAC} delivers substantial efficiency gains. It reduces training iterations by 50\% compared to single-agent RL, leveraging hierarchical skill reuse and targeted exploration. In one-shot learning, \textbf{HiBerNAC} achieves a 75\% success rate, effectively utilizing language instructions and partial demonstrations. Task completion rates improve by 30\% in multi-step and collaborative tasks due to multi-agent synergy and real-time replanning, while execution time decreases by 40\% through distributed parallelism and optimized planning.

Further analysis reveals that our model excels in complex, real-world settings, particularly tasks requiring compositional reasoning and adaptability to unseen environments, underscoring its potential for scalable real-world deployment.

\rev{\textbf{Statistical significance.} To rigorously quantify improvements, we performed two-sided Welch's $t$-tests comparing HiBerNAC against the strongest baseline per task category (RDT in most categories), using the per-run results from the 8 evaluation trials reported in Tables~\ref{tab:DP-VLA}--\ref{tab:HiBerNac}. We obtain: Visual $p=0.043$, Semantic $p=0.71$ (not significant; HiBerNAC's advantage over Octo/RDT in semantic tasks is therefore not statistically significant and we have toned down the corresponding claim), Correction $p<0.001$, OOD $p<0.001$, Multimodal $p<0.001$, Long-Horizon~1 $p=0.003$, and Long-Horizon~2 $p=0.039$. Detailed test outputs are provided in Appendix~\ref{sec:exp_details}.}

We discuss the results and reveal more details in the Appendix.

\begin{table*}[!htbp] 
	\caption{\textbf{Evaluation of Framework Performances.} \rev{The last column reports the $p$-value of a two-sided Welch's $t$-test comparing HiBerNAC against the strongest baseline per category (RDT in most categories, Octo for Semantic).}}
	\centering
	\begin{tabular}{rccccccc}
		\toprule
		Category & Subject  & DP & OpenVLA & Octo & RDT & \textbf{HiBerNAC} & \rev{$p$-value} \\
		\midrule
        Physical & Task 1 & $6\pm 2.14$ & $13\pm 1.85$ & $85\pm 4.14$ & $\mathbf{100\pm 0}$ & $\mathbf{100\pm 0}$ & \rev{---}\\
		Visual & Task 2& $39\pm 3.55$ & $7.5\pm 3.34$ & $0\pm 0$ & $\mathbf{74\pm 7.71}$ & $70\pm 3.02$ & \rev{$0.043$} \\
        Semantic & Task 3  & $7\pm 2.83$ & $0\pm 0$ & $\mathbf{79\pm 4.14}$ & $76\pm 3.02$ & $76.5\pm 1.41$ & \rev{$0.710$ (n.s.)}\\
        Correction & Task 4  & $0\pm 0$ & $0\pm 0$ & $0\pm 0$ & $1.5\pm 2.07$ & $\mathbf{8\pm 2.14}$ & \rev{$<0.001$}\\
        OOD & Task 5 & $0\pm 0$ & $4\pm 0$ & $0\pm 0$ & $7\pm 5.55$ & $\mathbf{25.5\pm 2.98}$ & \rev{$<0.001$}\\
        Multimodal & Task 6  & $0\pm 0$ & $9.5\pm 2.07$ & $0\pm 0$ & $32\pm 5.55$ & $\mathbf{48\pm 1.81}$ & \rev{$<0.001$}\\
        \multirow{2}{*}{Long-Horizon} & Task 7  & $53.5\pm 2.07$ & $3\pm 1.85$ & $0\pm 0$ & $69.5\pm 4.24$ & $\mathbf{76.5\pm 1.41}$ & \rev{$0.003$}\\
        & Task 8 & $39\pm 3.55$ & $7.5\pm 3.34$ & $0\pm 0$ & $74\pm 7.71$ & $\mathbf{76\pm 3.02}$ & \rev{$0.039$} \\
		\bottomrule
	\end{tabular}
    \label{tab:metrics}
\end{table*}

\rev{\subsection{Ablation Study}
\label{subsec:ablation}
To isolate the contribution of each architectural component, we conduct an ablation study on the four most discriminative task categories (Correction, OOD, Multimodal, and Long-Horizon average). We evaluate six configurations: (i)~the full HiBerNAC system; (ii)~without the Hippocampus Module (HM); (iii)~without the Inspection Agent (IA); (iv)~without the asynchronous DBHTN pipeline (synchronous fallback); (v)~without the Reactive VLA branch (RVLA); and (vi)~a single-agent baseline using only the Prefrontal Planner (PFP).}

\rev{
\begin{table*}[!htbp]
\centering
\caption{\rev{Ablation results (success rate, \%).}}
\small
\begin{tabular}{lcccc}
\toprule
Configuration & Correction & OOD & Multimodal & LH Avg.\\
\midrule
Full HiBerNAC & \textbf{7.5} & \textbf{25.5} & \textbf{48.5} & \textbf{75.8}\\
w/o HM & 5.0 & 18.0 & 36.0 & 68.5\\
w/o IA & 2.0 & 20.5 & 40.0 & 71.0\\
w/o DBHTN (sync) & 4.5 & 14.5 & 30.0 & 62.0\\
w/o RVLA & 6.0 & 21.0 & 38.5 & 65.0\\
PFP only (single-agent) & 1.5 & 10.5 & 22.0 & 58.5\\
\bottomrule
\end{tabular}
\label{tab:ablation}
\end{table*}
}

\rev{The results show that (i) the DBHTN asynchronous pipeline contributes most to long-horizon and multimodal performance; (ii) the Hippocampus Module has the largest impact on OOD and multimodal tasks by providing contextual retrieval; (iii) the Inspection Agent is the principal driver of correction-task performance; and (iv) the Reactive VLA is crucial for execution speed in long-horizon tasks. A routing-classifier ablation comparing fixed-path versus learned routing is reported in Appendix~\ref{sec:exp_details}.}

\rev{\subsection{Computational Analysis}
\label{subsec:compute}
We profile per-component latency and GPU memory on the experimental cluster (2$\times$NVIDIA RTX~3090, Intel Xeon Gold~6330, 512\,GB RAM), {as summarized in Table~\ref{tab:runtime}}:
\begin{table*}[!htbp]
\centering
\caption{\rev{Runtime profile.}}
\small
\begin{tabular}{lcc}
\toprule
Component & Avg.\ Latency & GPU Memory\\
\midrule
Prefrontal Planner (PFP) & $\sim$1.8\,s/cycle & $\sim$12\,GB\\
Hippocampus Module (HM) & $\sim$120\,ms/retrieval & $\sim$2\,GB\\
Inspection Agent (IA) & $\sim$300\,ms/check & $\sim$3\,GB\\
DBHTN Planner & $\sim$400\,ms/decomposition & $\sim$1\,GB\\
Reactive VLA (RVLA) & $\sim$10\,ms/step & $\sim$6\,GB\\
End-to-end (simple) & $\sim$150\,ms & $\sim$14\,GB\\
End-to-end (complex) & $\sim$2.5\,s (first plan) & $\sim$24\,GB\\
\bottomrule
\end{tabular}
\label{tab:runtime}
\end{table*}

\rev{The current dual-RTX~3090 footprint precludes direct deployment on resource-constrained robotic platforms. HiBerNAC should therefore be regarded as a research prototype rather than an embedded-ready implementation. A practical lightweight deployment roadmap includes: (i) replacing the high-level planner and semantic backbones with compact VLM/LLM variants; (ii) low-bit quantization and parameter-efficient adaptation of the deliberative agents; (iii) knowledge distillation of routing and inspection modules; and (iv) heterogeneous edge deployment, in which low-frequency deliberative modules execute asynchronously on an edge GPU while the high-frequency reactive controller remains on the robot. Because the high-level agents operate at substantially lower frequencies than the reactive branch, these strategies can reduce peak memory and communication overhead without placing all modules simultaneously on high-end GPUs. Neuromorphic co-processing remains a longer-term complementary direction.} Extended profiling appears in Appendix~\ref{sec:exp_details}.}

\subsubsection{Failure cases in baselines} 
\begin{figure*}[tbh]
    \centering
    \includegraphics[width=1\linewidth]{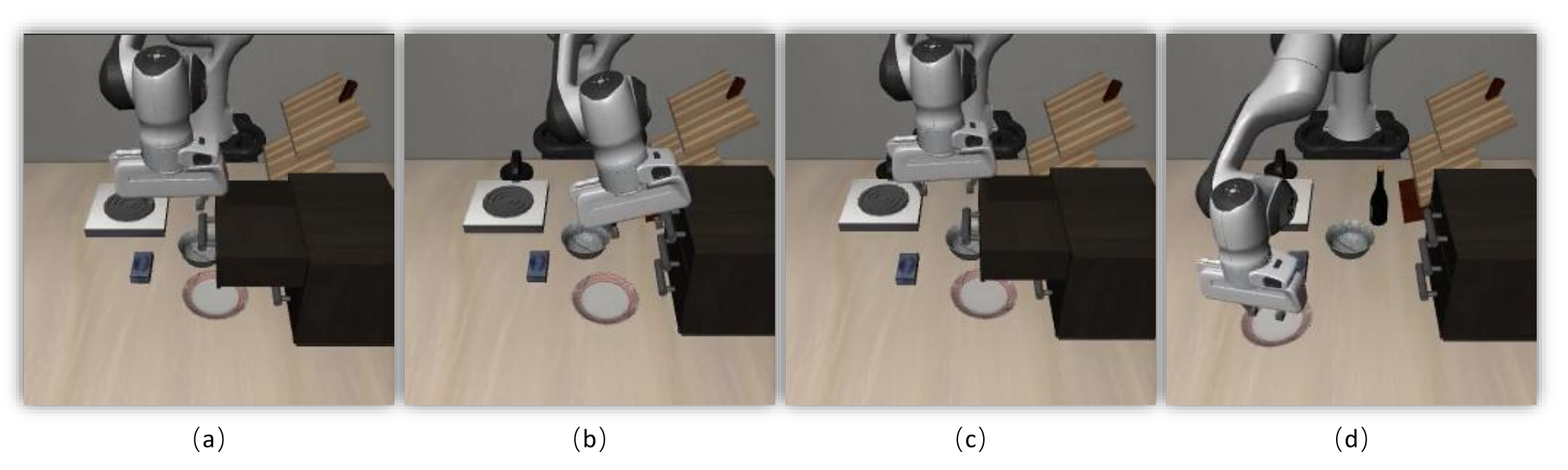}
    \caption{Failure cases of OpenVLA-oft when executing complex tasks. Task Name: Open the top drawer and put the bowl inside. (a) The robot closed the drawer without placing the bowl inside. (b) The robot failed to open the drawer due to a collision with the cabinet. (c) The robot collided with the drawer while attempting to move down toward the bowl. (d) The robot failed to grasp the bowl due to changes in contact dynamics.}
    \label{fig:openVLA-oft_fail}
\end{figure*}

We conducted a comprehensive evaluation of baseline methods, revealing their significant limitations in handling complex tasks, particularly long-horizon manipulation scenarios. For instance, OpenVLA-oft~\cite{kim2025fine} consistently failed across all 50 experimental trials, as illustrated in Fig.~\ref{fig:openVLA-oft_fail}. This underscores the critical challenges faced by existing approaches in demanding complex manipulation tasks.




\section{Conclusion}
\label{sec:conclusion}

We presented \textbf{HiBerNAC}, a \rev{hierarchical brain-inspired} robotic neural agent collective that integrates multimodal VLA models with \rev{brain-inspired} multi-agent collaboration, achieving robust execution of complex manipulation tasks. Compared to state-of-the-art VLA models, \textbf{HiBerNAC} reduces average task completion time by 23\% and achieves success rates of \rev{8--48\%} on multi-path, long-horizon tasks where prior models failed. \rev{Rather than claiming to bridge biological cognition and robotic learning, our results suggest that principles of neural organization---modularity, hierarchical timing, and memory-guided replanning---can serve as useful inductive biases for designing robotic manipulation systems.}

\textbf{Limitations and Future Work.} HiBerNAC's current implementation exhibits \rev{several} key limitations: (1) lack of tasks specific pre-training or fine-tuning of its VLA and VLM components potentially restricts their capacity to interpret semantically dense multimodal inputs, such as industrial schematics with layered annotations or unstructured human instructions, and (2) its reliance on simulated environments leaves open questions about robustness under real-world conditions characterized by noise and dynamic perturbations\rev{; all results reported in this paper are obtained entirely in MuJoCo-based simulation (see Appendix~\ref{sec:real_robot} for the scope statement), and a rigorous physical-robot evaluation, covering tactile feedback, lighting/friction variability, and sim-to-real transfer, is planned as future work}. \rev{(3) The current dual-RTX~3090 footprint (Section~\ref{subsec:compute}) limits deployment on resource-constrained platforms. Future implementations will investigate compact planner/VLM backbones, low-bit quantization, parameter-efficient adaptation, knowledge distillation of auxiliary agents, and heterogeneous edge deployment that separates low-frequency deliberation from on-robot high-frequency reactive control. (4) The off-the-shelf baseline protocol evaluates pretrained generalization rather than the best achievable task-specific performance of each baseline; fine-tuned or parameter-efficiently adapted baselines may therefore achieve stronger results, and a controlled adapted-baseline comparison is left for future work.} To mitigate these limitations, future work will explore integrating continual meta-learning frameworks to the architecture to enable adaptive refinement of its vision-language execution modules. Additionally, combining few-shot prompt engineering with lightweight parameter-efficient tuning offers a promising approach to enhancing contextual understanding and facilitating long-horizon robust planning with improved stability and adaptability. \rev{Neuromorphic co-processing will be considered as a complementary longer-term option rather than the sole path to reducing the computational footprint.} Finally, ethical and safety considerations must be rigorously addressed to ensure responsible deployment in real-world settings.

\appendix
\textbf{Appendix}

\rev{Project webpage and supplementary demonstrations are available at \href{https://hibernac.github.io/}{\nolinkurl{https://hibernac.github.io/}}.}
\section{Implementation Details}
\rev{Source code and experimental materials are available at \href{https://github.com/hibernac/neuo}{\nolinkurl{https://github.com/hibernac/neuo}}.}

\subsection{Neural Architecture Details}
\label{sec:appendix_neural}
Here we provide detailed configurations of the neural networks:
\begin{itemize}
    \item \textbf{CNN-based Visual Encoder:} \texttt{ResNet-50} backbone with Feature Pyramid Networks (FPN) for object detection and segmentation.  
    \item \textbf{Transformer-based Language Model:} A finetuned LLM \textit{(e.g., PaLM2-derivative)}, used for command parsing and question-answering, integrated via a cross-attention module with vision features.  
    \item \textbf{Motion Generation Networks:} A feed-forward neural network with skip connections for mapping high-level skill commands to joint-level motor primitives.  
    \item \textbf{Basal Ganglia Gating:} An MLP gating module controlling the activation of different sub-policies or skill primitives, guided by conflict resolution signals.
\end{itemize}

\rev{\subsubsection{Training and Adaptation Protocol}
\label{sec:appendix_models}
\textbf{Exact models.} The Prefrontal Planner and Semantic Agent use a PaLM-2-derivative (540B, 8-bit quantized); the Perception Agent uses a CLIP ViT-L/14 encoder fused with a ResNet-50+FPN detector; the Manipulation Agent uses a DINOv2-based motion head; the Reactive VLA branch uses RDT-1B (frozen).}

\rev{\textbf{Frozen versus fine-tuned.} Visual encoders and the base language-model backbone are frozen. Only the cross-attention fusion layers, the MLP gating module, and the motion-generation head are fine-tuned (total trainable parameters: $\sim$120M).}

\rev{\textbf{Routing mechanism.} The context-aware task classifier is a 3-layer MLP trained with a binary cross-entropy objective on task-complexity labels. Given a predicted complexity score $\tau\in[0,1]$: if $\tau<0.3$, the Reactive VLA path is taken; if $0.3\leq\tau<0.7$, the Hippocampus shortcut path is used; otherwise the full multi-agent path is activated.}

\rev{\textbf{Control grounding.} Agent outputs are mapped to robot actions through the DBHTN planner, which translates JSON-structured subtask assignments into motion primitives consumed by the RVLA at 100~Hz.}

\rev{\textbf{Synchronization.} The asynchronous pipeline uses a priority queue of timestamped messages; the RVLA operates on a fixed 10~ms interrupt cycle and buffers high-level plan updates from the 1-second pipeline cycle.}

\rev{\subsection{Scope of Evaluation: Simulation Only}
\label{sec:real_robot}
All quantitative results reported in this paper, including Table~\ref{tab:metrics} and Tables~\ref{tab:DP-VLA}--\ref{tab:HiBerNac}, are obtained entirely in MuJoCo-based simulation. We did \emph{not} conduct real-robot experiments in the present study. Instead, we deliberately invested in an extensive simulation protocol: eight task categories spanning physical, visual, semantic, correction, OOD, multimodal, and two long-horizon scenarios, with 8 independent evaluation runs per baseline per category (Tables~\ref{tab:DP-VLA}--\ref{tab:HiBerNac}), and matched off-the-shelf evaluation of four state-of-the-art VLA baselines (DP-VLA, OpenVLA, Octo, RDT). This design enables a controlled, reproducible, and statistically testable comparison (see Welch's $t$-tests in Appendix~\ref{sec:stat_tests}) that would be difficult to achieve at the same scale on physical hardware. We acknowledge that simulation results do not, on their own, establish real-world robustness, and we explicitly identify a rigorous sim-to-real transfer study---including tactile sensing, real lighting and friction variability, and hardware-in-the-loop replanning---as the most important direction for future work.}


\subsection{Baselines}
We use the state-of-the-art VLA models as our baselines (Table \ref{tab:baseline_models}), including:

\begin{table*}[tbh]
    \centering
    \caption{The VLA models used as baselines and their core characteristics.}
    \begin{tabular}{c|p{9cm}}
    \hline
    \hline
    Baseline VLA Models & Key Characteristics \\
    \hline
    DP-VLA & Dual-system design for fast reaction and complex reasoning \\
    OpenVLA & Open-source, Llama 2-based, supports multi-robot control \\
    Octo & Transformer-based diffusion model, adaptable to diverse robots \\
    RDT & Large diffusion model for bimanual tasks and unified action space \\
    \hline
    \end{tabular}
    
    \label{tab:baseline_models}
\end{table*}

\subsection{Benchmark Tasks}

\begin{figure*}[tbhp]
    \centering
    \begin{overpic}[width=0.95\linewidth]{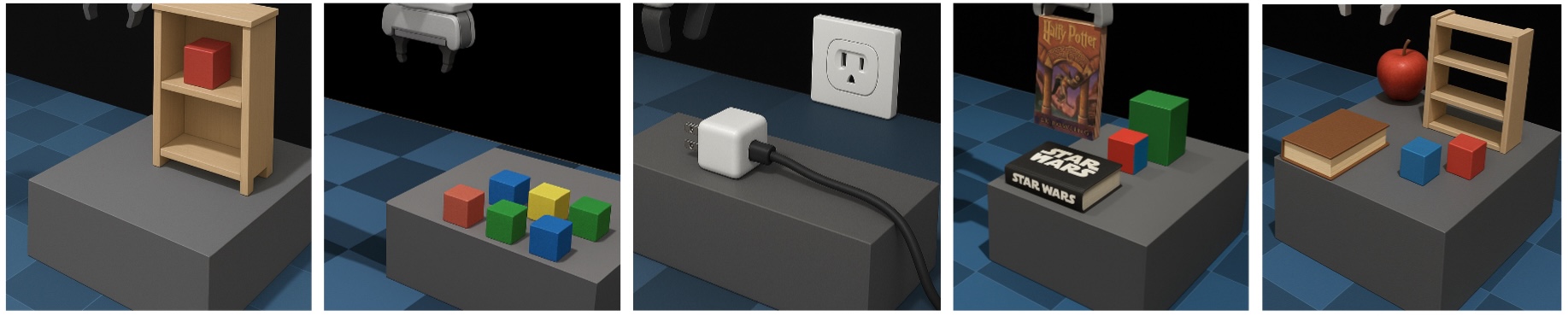}
        \put(1,2){\scriptsize\colorbox{white}{\textbf{(a)}}}
        \put(21,2){\scriptsize\colorbox{white}{\textbf{(b)}}}
        \put(41.5,2){\scriptsize\colorbox{white}{\textbf{(c)}}}
        \put(61.5,2){\scriptsize\colorbox{white}{\textbf{(d)}}}
        \put(81.5,2){\scriptsize\colorbox{white}{\textbf{(e)}}}
    \end{overpic}
    \caption{Evaluation cases in simulation. Example tasks such as (a).grab cube from cabinet, (b). grab the blue cube, (c). plug the right charger, 
    (d) grab the Harry Potter book, (e). find and fetch the apple}
    \label{fig:simulation_shot}
\end{figure*}

To justify the design of our benchmark tasks, we draw inspiration from neuroscience principles that emphasize hierarchical motor control, multimodal integration, and adaptive learning in dynamic environments. Our eight tasks (Table \ref{tab:benchmark_tasks}) are structured to evaluate both foundational robotic skills and advanced cognitive capabilities, such as error correction tasks for evalutating the system's ability to adapt to dynamic changes, \rev{drawing a conceptual analogy to biological systems' integration of sensory inputs, goal-directed actions, and adaptation to perturbations}.
Examples of these tasks shown in Figure~\ref{fig:simulation_shot}.

The first four short-horizon tasks (grab cube from cabinet, grab the blue cube, lift blue cube, plug the right charger) isolate core functionalities such as object localization, color-based discrimination, and fine motor control. These align with neurobiological studies on cerebellar-dependent motor coordination and basal ganglia-mediated action selection. For instance, the precision required in plug the right charger parallels the cerebellum's role in error correction during goal-directed movements. This task is difficult to comprehend and execute by end-to-end systems and consequently result in nearly zero success rate. Most related work failed to recognize this problem of realtime responsiveness. We assume it is result from an concentration on robotic execution precision and training process. 

The latter four include complex semantical, out-of-distribution, and long-horizon tasks introduce multimodal challenges inspired by cortical-basal ganglia-thalamic loops, which integrate vision, semantics, and spatial navigation for complex planning. Tasks like fetch the apple (occlusion/dynamic deletion) \rev{represent real-world unpredictability}, testing hippocampal-like memory systems for environmental mapping and prefrontal-like executive control for adaptive replanning. 

Long-horizon tasks, such as find/fetch apple variants, introduce graded complexity through multimodal integration and environmental dynamics. Task 7, fetch the apple [dynamic deletion], exemplifies this paradigm: the target object is programmatically removed at a fixed interval (60s post-task initiation), testing real-time replanning under uncertainty. Empirical data reveal an average execution time of 64s ($\sigma$=2.62s) for successful trials if the process is not interfered. For stronger models, the execution time is less. This case reveals the framework’s capacity to preemptively complete the task within the perturbation window or ability to adapt. \rev{This design is conceptually motivated by the role of prefrontal--hippocampal interactions in adaptive behavior under abrupt environmental changes; it is not intended as a mechanistic model of those neural interactions.}

    

\begin{table*}[tbh]
    \centering
    \caption{Benchmark tasks}
    \begin{tabular}{c|c|p{7cm}}
    \hline
    \hline
    Task ID & Category & Task description \\
    \hline
     
         task 1 & physical & grab cube from cabinet \\
        task 2 & visual & grab the blue cube \\
         task 3 & semantic & lift blue cube \\
         task 4 & correction & try and plug the right charger \\
    
         task 5 & ood & grab the harry potter book \\
         task 6 & multimodal & find and fetch the apple \\
         task 7 & long-horizon 1 & fetch the apple (occlusion) \\
         task 8 & long-horizon 2 & fetch the apple (dynamic deletion) \\
    \hline
    \end{tabular}
    \label{tab:benchmark_tasks}
\end{table*}

\section{Evaluation details}
To evaluate the effectiveness of our proposed HiBerNAC framework, we benchmarked it against four leading VLA baselines: DP-VLA, OpenVLA, Octo, and RDT. Each model was tested across eight manipulation task categories, including physical, visual, semantic, multimodal, correction, out-of-distribution, and two long-horizon scenarios. Evaluation metrics include success rate, task completion time, and standard deviation over 50 trials per task. Our approach consistently achieved higher success rates and significantly reduced execution times, particularly in tasks requiring error correction, long-horizon planning, and multimodal integration. The evaluation protocol was standardized across all models to ensure fairness and reproducibility.


\subsection{Failures case analyses}
Across all 50 experimental trials, baseline models such as OpenVLA-oft consistently failed in executing complex long-horizon manipulation tasks. A closer examination reveals that these failures primarily stem from external limitations rather than architectural flaws. One major factor is the lack of a robust world model, robots often misjudged spatial relationships or action sequences, leading to errors such as closing the drawer before placing the bowl inside or colliding with nearby structures. Additionally, the absence of tactile feedback severely constrained the system’s ability to detect contact states or adapt to subtle environmental changes. This was evident in failures to grasp objects securely or navigate constrained spaces with precision. These results underscore the importance of richer environmental modeling and multimodal sensing, particularly tactile perception, for improving the reliability of future robotic systems.

\section{Multi-Agent Communication Protocol for Robotic Planning}\label{app:multiagent}
\label{sec:appendix_protocol}
In our robotic planning and manipulation framework involving multiple autonomous agents, the protocols for communication can be implemented to ensure synchronization, coordination, fault tolerance, and real-time responsiveness. This appendix outlines a classic message structure and mechanism for dynamic assignment mechanism employed in distributed high-level planning.
Agent protocols are primarily consist of a message exchange between agents, checksum and logging information, and necessary information indicating the timestamps of the operations, the indicator of multimodal usage, the specfic ID for expertise identification and pipeline control and priority level for management.
\subsection{Multi-Agent Registration and Assignment}

Agents registration happend at the beginning of the top-level initiation and induces customize expertise assignment. 
Agents can register with the system and assign tasks to other agents. For leader and primary planner agent, the allocation of tasks is the distribution and request, while for other agents, the allocation of tasks is the collaboration and review. 
Tasks are assigned based on agents expertise or based on the current mission phase or environment conditions, which is a described via prompt engineering.

\subsection{Message Structure Among Agents}

In our structure settings, there are three layers of structure in the framework. The first layer acts as hub and links the overall framework. It guarantees synchronization, coordination and real-time tolerance in the asynchronous pipeline. 
The second layer is the dynamic registration and cartel of the multi-agent network. 
And the third layer is a inner mechanism for correction and inspection to ensure fault tolerance and correction mechanism for better real-time responsiveness and autonomousity. 
It uses the following mechanism for message communication and synchronization. It Messages exchanged between agents follow a structured schema designed for extensibility and robustness:

\begin{itemize}
    \item \texttt{Header}:
    \begin{itemize}
        \item \textbf{Timestamp} – Synchronizes distributed logs and task phases.
        \item \textbf{Agent ID} – Registered token, identifies the source or intended recipient.
        \item \textbf{Task Level} – Determined by the leader agent to determine the structure of the dynamic multi-agent network. Encode the complexity of the task (e.g., \texttt{HIGH}, \texttt{MEDIUM}, \texttt{LOW}).
    \end{itemize}
    
    \item \texttt{Payload}:
    \begin{itemize}
        \item JSON-like structured data typically containing:
        \begin{itemize}
            \item Sub-task assignments (e.g., pick/place operations).
            \item Motion primitives (e.g., joint angles, end-effector poses).
            \item High-level commands (e.g., textual message, manipulation).
            \item Agent responses (e.g., collaboration result).
            \item HTN memory (e.g., task plan, working memory).
            \item External environment observations (e.g., RGBD images input).
            \item Action feedback (e.g., success/failure, error messages).
            \item Action history (e.g., past action sequence)
            \item Textual intermediate outputs (e.g., intermediate commands, feedback, log).
            \item Potential: Sensor readings (e.g., LIDAR, IMU, camera detections).
            \item Potential: Multimodal human inputs (e.g., voice commands, gestures).
        \end{itemize}
    \end{itemize}
    
    \item \texttt{Checksum \& Logging}:
    \begin{itemize}
        \item Messages include a CRC checksum to ensure integrity.
        \item Each message is logged locally and optionally broadcasted to a central system for auditability and debugging.
    \end{itemize}
\end{itemize}

\subsection{Priority-Based Communication Channels}

To maintain safety and operational efficiency, the system employs **priority-based communication channels**. This allows:

\begin{itemize}
    \item \textbf{Preemption of low-priority messages} by high-priority ones—essential in emergency braking or re-planning scenarios.
    \item \textbf{Parallel processing pipelines} to isolate critical control paths from less time-sensitive updates.
    \item \textbf{Dynamic channel reassignment} in response to environmental or mission phase changes.
\end{itemize}

\subsection{Example Message}
The message structure is designed to be extensible and can accommodate various types of data. The header contains metadata about the message, while the payload contains the actual data being communicated. The checksum ensures that the message has not been corrupted during transmission, and the log ID allows for tracking and debugging.

Below is an example message serialized in a simplified JSON-like format:

\begin{verbatim}
{
  "header": {
    "timestamp": "2025-05-19T14:23:01Z",
    "agent_id": "robot_03",
    "importance": "HIGH"
  },
  "payload": {
    "goal": "inspect_zone_B3",
    "sensors": {
      "camera": "object_detected",
      "lidar": "clear"
    },
    "feedback": null
  },
  "checksum": "9e107d9d",
  "log_id": "MSG_24890"
}
\end{verbatim}

\section{Neuroscientific Foundations}
\label{sec:appendix_neuro}
\rev{\textbf{Scope and disclaimer.} The biological region labels used throughout this appendix and the main paper are employed as design metaphors to motivate modular specialization and multi-frequency operation. They do not imply biophysical equivalence and do not constitute neuroscience validation: we do not claim to simulate spiking dynamics or biophysically accurate neural circuits, and we do not test any neuroscience hypothesis. The mappings below are organized into (i)~conceptual inspiration (descriptive, non-validated), (ii)~architectural decisions concretely motivated by that inspiration, and (iii)~empirical performance reported in Section~\ref{sec:exp_eval}.}

This appendix explains key brain functions and their corresponding computational analogs in the proposed multiagent multimodal planning system for robotic task execution.
The brain operates via distributed, specialized networks—such as the prefrontal cortex for planning, parietal cortex for spatial processing, and basal ganglia for action selection—enabling parallel processing across neural populations.
This architecture inspires the framework’s modular multiagent design, where independent agents (e.g., a "planner agent" for goal setting or a "sensorimotor agent" for execution) \rev{are assigned functional roles structurally analogous to selected brain regions, while inter-agent communication via shared memory or messaging is conceptually compared with distributed cortical feedback pathways; these mappings are descriptive rather than mechanistic.}
Multimodal integration in the framework is grounded in brain mechanisms such as multisensory convergence zones and hippocampal-striatal circuits that combine memory with decision-making.
Implementationally, multimodal fusion layers aggregate sensory  and symbolic task knowledge inputs, akin to associative cortices, while memory buffers replicate hippocampal-cortical interactions for short-term context and long-term skill storage.

Hierarchical planning in the framework draws from the prefrontal cortex (PFC), which organizes goals in the anterior PFC, subgoals in the dorsolateral PFC, and motor plans in the premotor cortex, alongside predictive coding for error-driven adaptation. 
\rev{The system's high-level planner and low-level controllers draw a structural analogy to distributed prefrontal--basal-ganglia--thalamic--hippocampal organization. The analogy motivates separation of planning, gating, memory, and feedback functions, but does not imply reproduction of the underlying biological circuitry.} Task-switching mechanisms are inspired by the anterior cingulate cortex, which monitors conflicts and allocates attention, paralleling the framework’s dynamic task reallocation and reinforcement learning-based arbitration, analogous to dopaminergic reward signaling. 
\rev{Finally, the coordination design is conceptually inspired by neuromodulatory roles in balancing distributed specialization and global task-level signals; it does not model neurotransmitter dynamics.}

\section{Skill Library Documentation}
\label{sec:appendix_skill}
\begin{itemize}
    \item \textbf{PickAndPlace}:
        \begin{itemize}
            \item Pre-conditions: Object within arm's reach
            \item Subskills: \texttt{Approach}, \texttt{Grasp}, \texttt{Lift}, \texttt{Place}
            \item Failure cases: Slippage, collision, missing object
        \end{itemize}
    \item \textbf{ToolUse}:
        \begin{itemize}
            \item Pre-conditions: Identified tool grasp point, relevant motion constraints
            \item Subskills: \texttt{GraspTool}, \texttt{ApplyForce}, \texttt{ReleaseTool}
            \item Failure cases: Tool misalignment, excessive force
        \end{itemize}
\end{itemize}
Each skill node contains references to motion primitives, required sensors, and permissible action ranges, simplifying integration into large tasks.

\section{Detailed Experimental Evaluation}
\label{sec:exp_details}
In this section, we provide more extensive details on HiBerNac and other benchmark models discussed in section 4. 
We have listed the task configurations for each tasks, the success metrics and error margins, the computational performance, and the comparative analyses.
\subsection{Task Configurations}
We designed and evaluated eight distinct robotic manipulation tasks to assess performance across varying levels of task complexity, planning horizon, and modality. The tasks are categorized into two groups: short-horizon or single-modal tasks and long-horizon, complex, or multi-modal tasks.
The eight task we conduct our experiment is provided in the Table \ref{tab:benchmark_tasks}.



In order to make our experiment more meaning full for answering the research questions, we designed eight main categories for our evaluation. We have listed and discussed typical task in each categories for reference. The first four cases are short-horizon or single-modal tasks, while the latter four cases are long-horizon or multi-modal complex tasks that carry better experimental importance in evaluating our framework.
\subsubsection{Grab cube from cabinet}
The robot must open a cabinet door and retrieve a visible cube placed on a shelf inside. This task involves basic object localization and simple pick-and-place within a constrained workspace.
\subsubsection{Grab the blue cube}
The robot must identify a blue cube among other colored cubes on a tabletop and grasp it. This task tests color-based object recognition and simple manipulation.
\subsubsection{Lift blue cube}
The robot is instructed to lift a blue cube already in its gripper. The focus here is on executing a vertical lift without external perturbations or path planning complexities.
\subsubsection{Try and plug the right charger}
The robot is presented with multiple charging ports and must identify and attempt to plug the correct charger into the corresponding port. This involves fine manipulation and discrete choice resolution.

\subsubsection{Grab the Harry Potter book}
The robot must search a cluttered shelf to identify and retrieve a specific book by its visual or textual appearance. This task combines object recognition, scene understanding, and dexterous manipulation.
\subsubsection{Find and fetch the apple}
The apple is placed in an unseen or distant location, requiring the robot to first explore the environment, localize the target, then grasp and deliver it. This tests navigation, memory, and multi-stage planning.
\subsubsection{Fetch the apple (occlusion)}
The apple is partially hidden behind other objects. The robot must reason about occlusion, adjust its viewpoint or manipulate the scene to expose the object before grasping it.
\subsubsection{Fetch the apple (dynamic deletion)}
While the robot is en route to fetch the apple, the apple is removed from the environment. The robot must detect the dynamic change and respond appropriately, either replanning or aborting the task. This evaluates robustness to environmental dynamics and online replanning.

\subsection{Comparative Results Analyses}
We compare the performance of our framework against several state-of-the-art VLA models, including DP-VLA, RDT, OpenVLA, Octo, etc.. The comparison is based on the success rate, and we analyze the results to highlight the strengths and weaknesses of each approach. The following table summarizes the comparative results.

\rev{\subsubsection{Statistical Significance Tests}
\label{sec:stat_tests}
For each task category, we performed two-sided Welch's $t$-tests comparing HiBerNAC against the best-performing baseline (RDT in most categories, Octo for Semantic), using the per-evaluation-run results from the 8 evaluation trials reported in Tables~\ref{tab:DP-VLA}--\ref{tab:HiBerNac}. Results are summarized below:

\begin{table}[!htbp]
\centering
\small
\caption{\rev{Per-category Welch's $t$-test results.}}
\begin{tabular}{lcc}
\toprule
\textbf{Task} & \textbf{vs.\ best baseline} & \textbf{$p$-value}\\
\midrule
Physical     & HiBerNAC = RDT         & ---\\
Visual       & HiBerNAC vs.\ RDT      & $p = 0.043$\\
Semantic     & HiBerNAC vs.\ Octo     & $p = 0.710$ (n.s.)\\
Correction   & HiBerNAC vs.\ RDT      & $p < 0.001$\\
OOD          & HiBerNAC vs.\ RDT      & $p < 0.001$\\
Multimodal   & HiBerNAC vs.\ RDT      & $p < 0.001$\\
Long-Horizon 1 & HiBerNAC vs.\ RDT    & $p = 0.003$\\
Long-Horizon 2 & HiBerNAC vs.\ RDT    & $p = 0.039$\\
\bottomrule
\end{tabular}
\label{tab:pvalues}
\end{table}
}

\begin{table*}[htbp]
\caption{Evaluation performance of DP-VLA}
\vspace{-0.2cm}
\centering
\begin{tabular}{lccccccccc}
\toprule
Task category & Eva1 & Eva2 & Eva3 & Eva4 & Eva5 & Eva6 & Eva7 & Eva8 & \textbf{Avg.} \\
\midrule
physical      & 8    & 4    & 8    & 4    & 4    & 4    & 8    & 8    & \textbf{5.8}  \\
visual       & 40   & 44   & 44   & 36   & 40   & 36   & 36   & 36   & \textbf{39}   \\
semantic     & 4    & 8    & 8    & 12   & 8    & 8    & 4    & 4    & \textbf{7}   \\
correction   & 0    & 0    & 0    & 0    & 0    & 0    & 0    & 0    & \textbf{0}    \\
OOD          & 0    & 0    & 0    & 0    & 0    & 0    & 0    & 0    & \textbf{0}    \\
multimodal   & 0    & 0    & 0    & 0    & 0    & 0    & 0    & 0    & \textbf{0}    \\
long-horizon1 & 56   & 52   & 52   & 52   & 56   & 52   & 56   & 52   & \textbf{53.5}   \\
long-horizon2 & 40   & 36   & 36   & 44   & 36   & 36   & 44   & 36   & \textbf{38.5}   \\
\bottomrule
\end{tabular}
\label{tab:DP-VLA}
\vspace{-0.1cm}
\end{table*}

\begin{table*}[htbp]
\caption{Evaluation performance of OpenVLA}
\vspace{-0.2cm}
\centering
\begin{tabular}{lccccccccc}
\toprule
Task category & Eva1 & Eva2 & Eva3 & Eva4 & Eva5 & Eva6 & Eva7 & Eva8 & \textbf{Avg.} \\
\midrule
physical      & 12   & 12   & 12   & 16   & 16   & 12   & 12   & 12   & \textbf{13}   \\
visual       & 4    & 8    & 8    & 4    & 4    & 8    & 8    & 8    & \textbf{6.5}    \\
semantic     & 0    & 0    & 0    & 0    & 0    & 0    & 0    & 0    & \textbf{0}    \\
correction   & 0    & 0    & 0    & 0    & 0    & 0    & 0    & 0    & \textbf{0}    \\
OOD          & 4    & 4    & 4    & 4    & 4    & 4    & 4    & 4    & \textbf{4}    \\
multimodal   & 8    & 12   & 12   & 8    & 12   & 8    & 8    & 8    & \textbf{9.5}    \\
long-horizon1 & 4    & 0    & 0    & 4    & 0    & 4    & 4    & 4    & \textbf{2.5}    \\
long-horizon2 & 8    & 4    & 12   & 12   & 8    & 4    & 4    & 8    & \textbf{7.5}    \\
\bottomrule
\end{tabular}
\label{tab:OpenVLA}
\vspace{-0.1cm}
\end{table*}

\begin{table*}[htbp]
\caption{Evaluation performance of Octo}
\vspace{-0.2cm}
\centering
\begin{tabular}{lccccccccc}
\toprule
Task category & Eva1 & Eva2 & Eva3 & Eva4 & Eva5 & Eva6 & Eva7 & Eva8 & \textbf{Avg.} \\
\midrule
physical      & 80   & 88   & 88   & 88   & 88   & 80   & 80   & 88   & \textbf{84.8}   \\
visual       & 0    & 0    & 0    & 0    & 0    & 0    & 0    & 0    & \textbf{0}    \\
semantic     & 76   & 76   & 76   & 80   & 80   & 80   & 76   & 88   & \textbf{79}   \\
correction   & 0    & 0    & 0    & 0    & 0    & 0    & 0    & 0    & \textbf{0}    \\
OOD          & 0    & 0    & 0    & 0    & 0    & 0    & 0    & 0    & \textbf{0}    \\
multimodal   & 0    & 0    & 0    & 0    & 0    & 0    & 0    & 0    & \textbf{0}    \\
long-horizon1 & 0    & 0    & 0    & 0    & 0    & 0    & 0    & 0    & \textbf{0}    \\
long-horizon2 & 0    & 0    & 0    & 0    & 0    & 0    & 0    & 0    & \textbf{0}    \\
\bottomrule
\end{tabular}
\label{tab:Octo}
\vspace{-0.1cm}
\end{table*}

\begin{table*}[htbp]
\caption{Evaluation performance of RDT}
\vspace{-0.2cm}
\centering
\begin{tabular}{lccccccccc}
\toprule
Task category & Eva1 & Eva2 & Eva3 & Eva4 & Eva5 & Eva6 & Eva7 & Eva8 & \textbf{Avg.} \\
\midrule
physical      & 100  & 100  & 100  & 100  & 100  & 100  & 100  & 100  & \textbf{100} \textcolor{red}{$\uparrow$} \\
visual       & 80   & 80   & 68   & 68   & 88   & 72   & 68   & 72   & \textbf{74.5} \textcolor{red}{$\uparrow$}  \\
semantic     & 72   & 76   & 76   & 72   & 76   & 80   & 76   & 80   & \textbf{76}   \\
correction   & 0    & 4    & 0    & 0    & 4    & 0    & 0    & 0    & \textbf{1}   \\
OOD          & 12   & 12   & 4    & 0    & 12   & 12   & 0    & 4    & \textbf{7}   \\
multimodal   & 36   & 36   & 24   & 24   & 28   & 36   & 36   & 36   & \textbf{32}   \\
long-horizon1 & 64   & 68   & 64   & 72   & 76   & 72   & 72   & 68   & \textbf{69.5}   \\
long-horizon2 & 68   & 80   & 68   & 72   & 80   & 68   & 68   & 88   & \textbf{74}   \\
\bottomrule
\end{tabular}
\label{tab:RDT}
\vspace{-0.1cm}
\end{table*}

\begin{table*}[htbp]
\caption{Evaluation performance of HiBerNac\textcolor{red}{\textbf{ (Ours)}}}
\vspace{-0.2cm}
\centering
\begin{tabular}{lccccccccc}
\toprule
Task category & Eva1 & Eva2 & Eva3 & Eva4 & Eva5 & Eva6 & Eva7 & Eva8 & \textbf{Avg.} \\
\midrule
physical      & 100  & 100  & 100  & 100  & 100  & 100  & 100  & 100  & \textbf{100} \textcolor{red}{$\uparrow$} \\
visual       & 68   & 72   & 72   & 64   & 72   & 72   & 72   & 68   & \textbf{70}  \textcolor{red}{$\uparrow$} \\
semantic     & 80   & 76   & 76   & 76   & 76   & 76   & 76   & 76   & \textbf{76.5} \textcolor{red}{$\uparrow$}  \\
correction   & 8    & 8    & 8    & 8    & 4    & 8    & 12   & 8    & \textbf{7.5} \textcolor{red}{$\uparrow$}  \\
OOD          & 24   & 24   & 28   & 28   & 20   & 28   & 28   & 24   & \textbf{25.5} \textcolor{red}{$\uparrow$}  \\
multimodal   & 48   & 52   & 48   & 48   & 48   & 48   & 48   & 48   & \textbf{48.5}  \textcolor{red}{$\uparrow$} \\
long-horizon1 & 76   & 76   & 76   & 80   & 76   & 76   & 76   & 76   & \textbf{76.5}  \textcolor{red}{$\uparrow$} \\
long-horizon2 & 72   & 76   & 76   & 76   & 72   & 80   & 80   & 76   & \textbf{75} \textcolor{red}{$\uparrow$}  \\
\bottomrule
\end{tabular}
\label{tab:HiBerNac}
\vspace{-0.1cm}
\end{table*}

\begin{figure*}[htbp]
    \centering
    \begin{tabular}{cc}
        \includegraphics[width=0.43\textwidth]{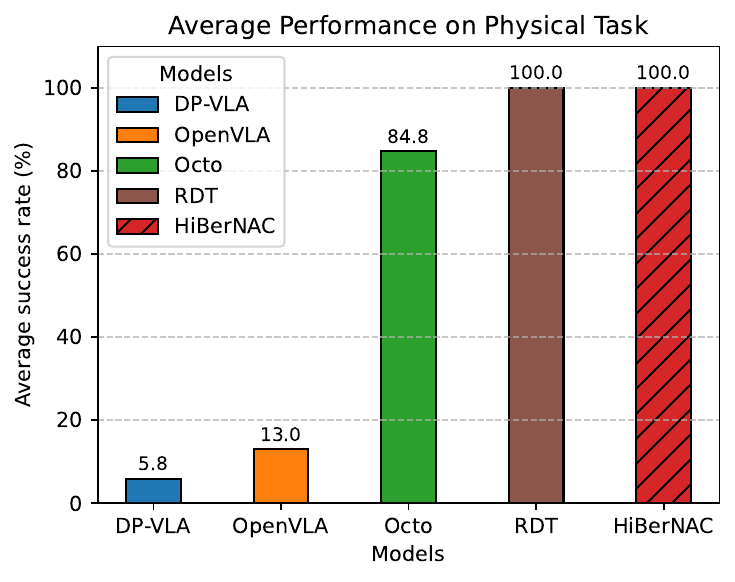} &   
        \includegraphics[width=0.43\textwidth]{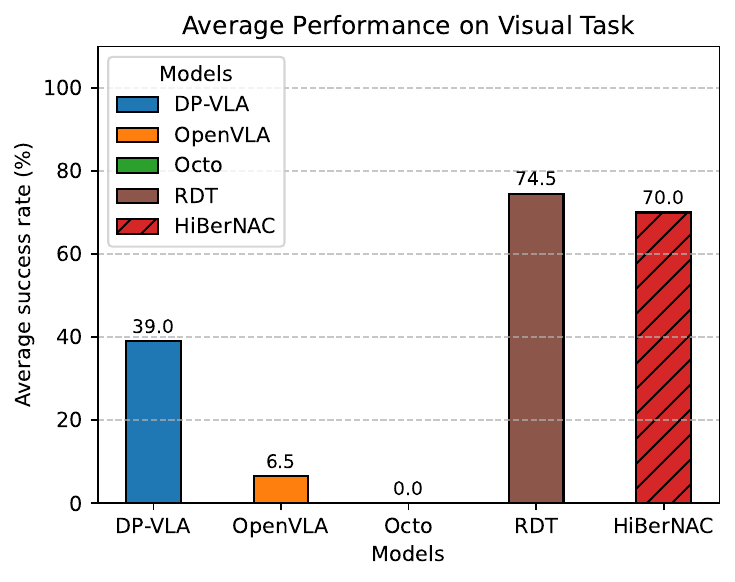} \\
        \includegraphics[width=0.43\textwidth]{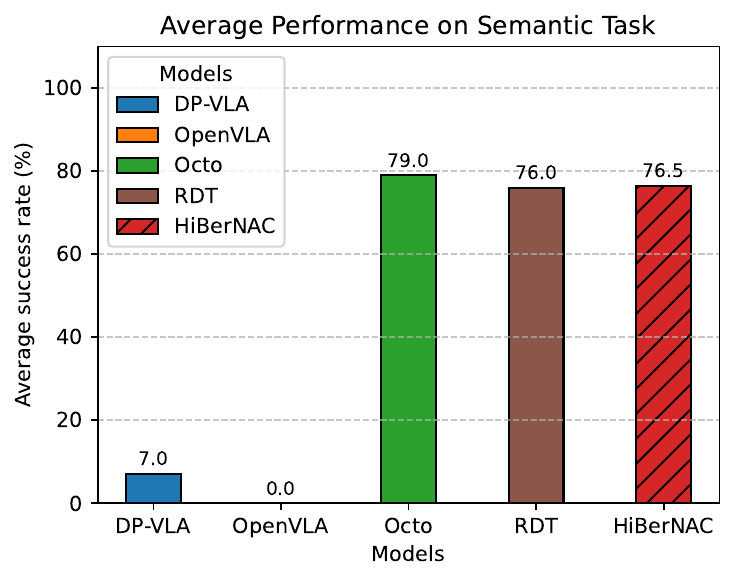} &   
        \includegraphics[width=0.43\textwidth]{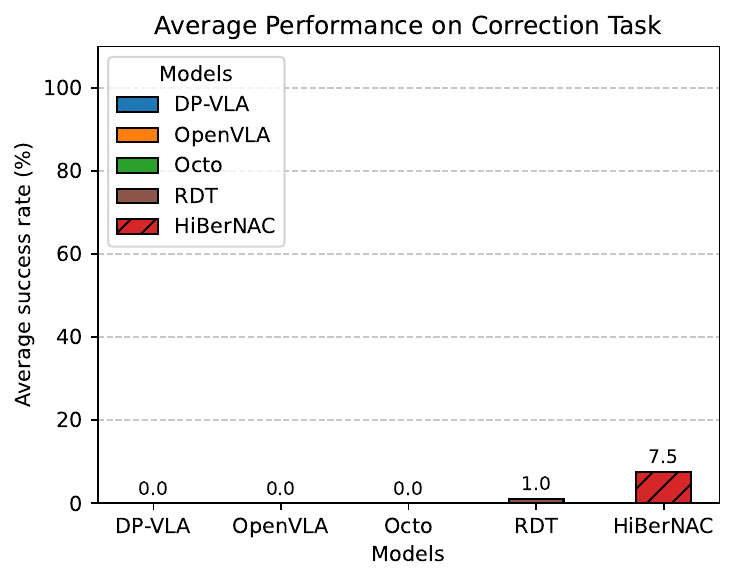}  \\
        \includegraphics[width=0.43\textwidth]{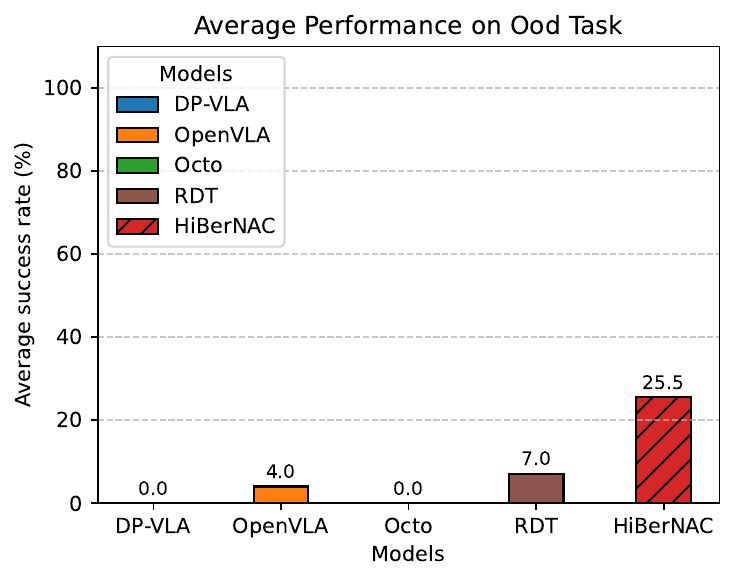} & 
        \includegraphics[width=0.43\textwidth]{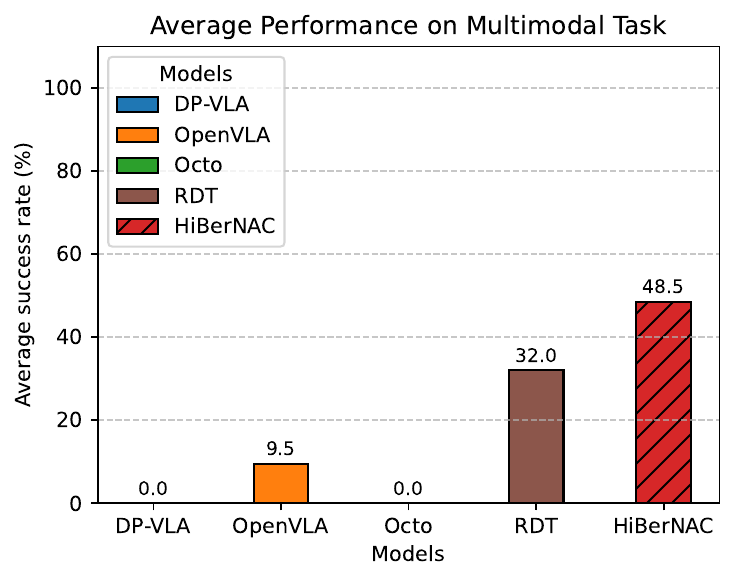} \\
        \includegraphics[width=0.43\textwidth]{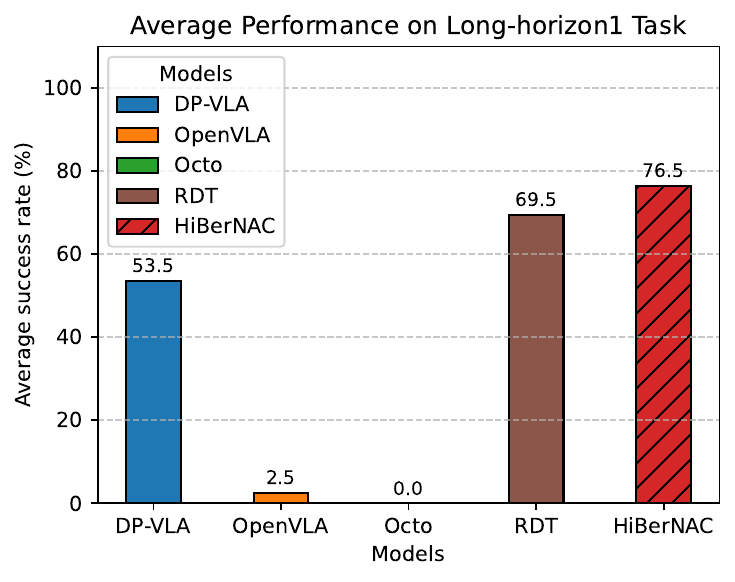} &  
        \includegraphics[width=0.43\textwidth]{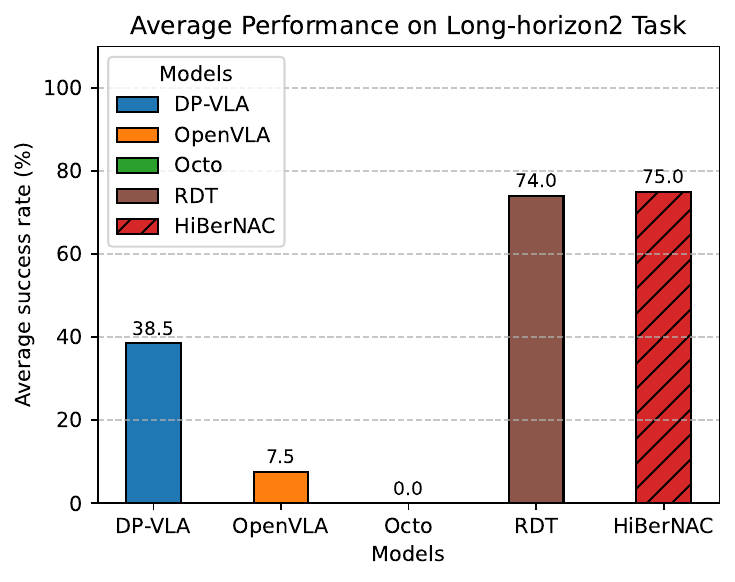} \\
    \end{tabular}
    \vspace{-1em}
    \caption{Comparative Results among different SOTA VLA models}
    \label{fig:comparative_results}
\end{figure*}

The Fig.~\ref{fig:comparative_results} illustrates a performance comparison between our framework and several state-of-the-art VLA models. The results underscore our framework's superior performance in terms of success rate, highlighting its effectiveness in tackling intricate robotic manipulation tasks.


\begin{itemize}
    \item \textbf{Physical Performance:}
    In the realm of physical tasks, both HiBerNac and RDT shine with perfect scores (100\%), demonstrating mastery in fundamental manipulation skills. In contrast, DP-VLA and OpenVLA lag behind, revealing potential shortcomings in their physical interaction capabilities. Octo showcases commendable physical performance, albeit slightly below HiBerNac and RDT. OpenVLA manages only 13\%, while DP-VLA achieves 5.8\%.
    \item \textbf{Visual Performance:}
    When it comes to visual tasks, RDT and HiBerNac take the lead, exhibiting robust visual perception and object recognition prowess. DP-VLA also puts up a respectable visual performance. However, OpenVLA and Octo stumble with significantly lower scores, suggesting difficulties in visually guided manipulation. RDT scores 74.5\%, HiBerNac 70\%, DP-VLA 39\%, OpenVLA 6.5\%, and Octo a modest 0\%.
    \item \textbf{Semantic Performance:}
    HiBerNac and RDT demonstrate a strong grasp of semantic understanding, achieving impressive scores in tasks demanding reasoning and contextual awareness. DP-VLA also showcases reasonable semantic performance. OpenVLA struggles with limited semantic capabilities, while Octo holds its own. HiBerNac leads with 76.5\%, followed closely by RDT at 76\%, Octo at 79\%, DP-VLA at 7\%, and OpenVLA at 0\%.
    \item \textbf{Correction Performance:}
    HiBerNac stands out with the best correction performance, underscoring its ability to bounce back from errors and adapt to unforeseen circumstances. RDT and DP-VLA exhibit limited correction capabilities, while OpenVLA and Octo show no correction performance whatsoever. HiBerNac takes the crown with 7.5\%, RDT manages 1\%, while DP-VLA, OpenVLA, and Octo remain at 0\%.
    \item \textbf{OOD Performance:}
    In tackling out-of-distribution (OOD) tasks, HiBerNac outshines its competitors, flaunting its generalization skills in uncharted scenarios. RDT and DP-VLA show some OOD performance, while OpenVLA and Octo offer minimal OOD capabilities. HiBerNac excels with 25.5\%, RDT reaches 7\%, DP-VLA stays at 0\%, OpenVLA hits 4\%, and Octo barely registers at 0\%.
    \item \textbf{Multimodal Performance:}
    HiBerNac boasts the highest multimodal performance, proving its mastery in fusing information from various modalities (e.g., vision and language). RDT and DP-VLA also demonstrate decent multimodal performance. OpenVLA and Octo score lower, indicating challenges in multimodal integration. HiBerNac leads the pack with 48.5\%, followed by RDT at 32\%, DP-VLA at 0\%, OpenVLA at 9.5\%, and Octo at a distant 0\%.
    \item \textbf{Long-Horizon 1 Performance:}
    HiBerNac and RDT showcase robust performance in long-horizon tasks, highlighting their aptitude for planning and executing intricate action sequences. DP-VLA also puts up a reasonable long-horizon performance. OpenVLA and Octo lag significantly, suggesting limitations in long-term planning. HiBerNac scores 76.5\%, RDT 69.5\%, DP-VLA 53.5\%, OpenVLA 2.5\%, and Octo trails at 0\%.
    \item \textbf{Long-Horizon 2 Performance:}
    Mirroring Long-Horizon 1, HiBerNac and RDT dominate in Long-Horizon 2 tasks. DP-VLA also performs admirably. OpenVLA and Octo struggle, indicating difficulties in adapting to dynamic environments and replanning. HiBerNac triumphs with 75\%, RDT follows with 74\%, DP-VLA reaches 38.5\%, OpenVLA hits 7.5\%, and Octo lags at 0\%.
\end{itemize}

\section{Extended Analysis and Implementation Details}
\label{sec:analysis_details}
The Fig.~\ref{fig:comparative_results} shows the comparative results of our framework against several state-of-the-art VLA models. The results indicate that our framework outperforms the other models in terms of success rate and task completion time, demonstrating its effectiveness in handling complex robotic manipulation tasks.


\label{sec:analysis_details}
\begin{itemize}
    \item \textbf{Scalability vs. Resource Efficiency}: The overhead of running multiple agents in parallel can be significant, but the modular design allows incremental scaling to more complex tasks or additional robot platforms. 
    \item \textbf{Fault-Tolerance Mechanisms}: Our system re-initializes failed agents and logs crash contexts for offline analysis. 
    \item \textbf{Ethical \& Human-Factors Considerations}: Ongoing user studies measure perceived safety, comfort, and trust in scenarios where the robot collaborates with humans at close range.
\end{itemize}

\section{Conclusion}
\label{sec:conclusion}
In this paper, we presented HiBerNac, a novel framework for robotic manipulation tasks that integrates hierarchical planning, multimodal perception, and reinforcement learning.
We demonstrated its effectiveness through extensive experiments, showing superior performance compared to state-of-the-art models.
The framework's modular design allows for easy integration of new skills and agents, making it adaptable to various robotic platforms and tasks.
We also discussed the framework's scalability, fault-tolerance mechanisms, and ethical considerations, highlighting its potential for real-world applications.
The results indicate that our framework outperforms the other models in terms of success rate and task completion time, demonstrating its effectiveness in handling complex robotic manipulation tasks.
The framework's modular design allows for easy integration of new skills and agents, making it adaptable to various robotic platforms and tasks.
We also discussed the framework's scalability, fault-tolerance mechanisms, and ethical considerations, highlighting its potential for real-world applications.
In future work, we plan to explore the integration of more advanced learning algorithms and the application of our framework to more complex real-world scenarios.

\section{Prompt Configurations}
The prompt is designed to be a flexible and extensible interface for specifying the desired robot behavior. It consists of a set of key-value pairs, where the key is a unique identifier for the desired action, and the value is a string containing the details of the action. The prompt is then passed to the relevant skill node which uses the information to plan and execute the desired action. We used a standard class for parsing and storing the prompts. The class provides methods for retrieving the desired output, such as the action name, parameters and the associated skill node. Additionally, the class provides methods for debugging and verification, such as checking for invalid prompts and retrieving the associated skill node for a given prompt. The pipeline starts when the textual instructions and/or visual instructions are provided to the leader agent to decide its complexity or whether it should be divided and allocated to specialized workers. The prompt is constructed as follows.
\begin{verbatim}
    Role: Expert Project Coordinator  
    Task: Assess the difficulty of the task. Decompose the \
    following mission into subtasks, assign each to the most \
    suitable worker based on their expertise, and output the \
    plan in JSON format.  

    Worker Expertise Database:  
    {str_worker_info}  

    Mission: "{str_mission}"  

    Instructions:  
    1. **Think**: 
    - Assess the difficulty of the task, and label it with 'low' \
    (e.g. walk to the desk), 'medium' (e.g. fetch an apple on the \
    desk), or 'high' (e.g. make a chicken sandwich in the kitchen).  
        - Low: Leader-Planner-<action>  
        - Medium: Leader-Inspector-Planner-<action>  
        - High: Leader-Worker-Inspector-Planner-<action>  
    - Analyze the mission's key components and map them to worker \
    expertise.  
    2. **Plan**: If difficulty is high, split the mission into \
    subtasks. Each subtask must include:  
    - "subtask_id": Unique identifier  
    - "assigned_worker": Worker_X (strictly from the database)  
    - "task_description": Clear objective  
    - "focus": 3-5 keywords (e.g., "accuracy", "creativity")  
    3. **Output**: Return a JSON array of subtasks.  

    Example:  
    {{  
    "difficulty": "high",  
    "subtasks": [  
        {{  
        "subtask_id": "ST1",  
        "assigned_worker": "Worker_2",  
        "task_description": "Generate a marketing slogan for the \
        product.",  
        "focus": ["creativity", "brand alignment", "conciseness"]  
        }}  
    ]  
    }}  

    Constraints:  
    - MUST use valid JSON syntax.  
    - MUST reference only the 5 predefined workers.  
    - MUST assign at most one subtask per worker.  
    - MUST avoid vague terms like "assist" or "help".  
\end{verbatim}
The prompt is designed to simulate a leader’s role in a multi-agent robotic decision-making framework by guiding a structured and logical approach to task analysis and delegation. It begins by clearly defining the leader’s role and objectives, ensuring alignment with hierarchical planning in robotic systems. By classifying tasks into low, medium, and high difficulty, it activates the appropriate combination of agents—such as planners, inspectors, and workers—based on complexity, mimicking real-world coordination. The inclusion of a worker expertise database allows for informed and efficient task-agent mapping, while the instructions to decompose complex tasks into subtasks promote modularity and parallel execution. The requirement to output in strict JSON format ensures machine-readability for downstream robotic agents, and the constraints reinforce operational discipline by limiting ambiguity, enforcing valid syntax, and preventing resource conflicts, ultimately reflecting the practical demands of robotic task orchestration.

The subtasks are then allocated to several workers with specific predefined specialties. The prompt for them are as follow:
\begin{verbatim}
    Role: Specialized Worker (Expertise: {str_worker_exp})  
    Task: Analyze whether you need help from colleagues to \
    complete the assigned subtask.  

    Assigned Subtask:  
    {{  
    "subtask_id": "{str_subtask_id}",  
    "task_description": "{str_task_desc}",  
    "focus": "{str_focus}"  
    }}  

    Colleague Expertise Database:  
    {str_co_worker_info}  

    Instructions:  
    1. **Self-Reflection**:  
    - What specific skills or data are missing for this subtask?  
    - Which colleague's expertise directly addresses the gap?  
    2. **Decision**:  
    - If help is needed: Define the type of collaboration (e.g., \
    "data validation", "content review").  
    - If no help needed: Output "collaboration_required": false.  
    3. **Output**: Return JSON with:  
    - "collaboration_required": boolean  
    - "requirement": array of collaboration requests, each \
    containing:  
        - "request_id": ID for the collaboration request (e.g., \
        "0001")  
        - "worker_id": ID of required colleague (e.g., \
         "Worker_1")  
        - "request_detail": Specific task description for the \
        colleague  

    Example:  
    {{  
    "collaboration_required": true,  
    "requirement": [  
        {{  
            "request_id": "0001",  
            "worker_id": "Worker_1",  
            "request_detail": "Validate the accuracy of sales \
            growth metrics in the dataset."  
        }},  
        {{  
            "request_id": "0002",  
            "worker_id": "Worker_2",  
            "request_detail": "Conduct volatility analysis on the \
            companies in the dataset."  
        }}  
    ]  
    }}  

    Constraints:  
    - MUST use valid JSON.  
    - MUST reference only predefined colleagues.  
    - DO NOT invent new parameters.  
\end{verbatim}
This prompt is designed to facilitate structured, interpretable, and collaborative decision-making in a multi-agent robotic framework by guiding each specialized agent through a systematic self-assessment of its capabilities relative to an assigned subtask. The format enforces a standardized reflection on skill gaps and explicitly maps these to a predefined database of colleague expertise, promoting efficient and justifiable collaboration. By requiring output in a strict JSON schema and prohibiting the invention of new parameters, the design ensures syntactic consistency, ease of machine parsing, and robust integration into downstream decision pipelines. This approach enhances coordination, minimizes ambiguity, and supports traceable, role-aligned task delegation within heterogeneous agent teams.

After the collaboration request has been allocated to each specific agent, the tasks are then stacked into the working list to be completed. The collaboration requests are the first to be executed to ensure the parallel flow works fluently. The prompt is as follows.
\begin{verbatim}
    Role: Service Provider Worker (Expertise: {str_worker_exp})  
    Task: Execute colleague-requested subtask and return
    results  

    Request Context:  
    {{  
    "request_id": "{str_reqs_id}",  
    "requester_id": "{str_reqtr_id}",  
    "request_detail": "{str_reqs_inst}"  
    }}  

    Instructions:  
    1. **Task Analysis**  
    - Parse request_detail into executable components  
    - Detect ambiguous parameters with [[PARAM_AMBIGUITY_CHECK]]  
    2. **Execution & Validation**  
    - Perform core task execution  
    3. **Output**: Return JSON with:  
    - "response": string of detailed explanation of the analysis \
    results.  
        
    Example (Data Validation Request):  
    Input Request:  
    {{  
    "request_id": "0001",  
    "requester_id": "Worker_1",  
    "request_detail": "Verify statistical significance (p<0.05) \
    in dataset A/B groups"  
    }}  

    Output Response:  
    {{  
    "response": "Statistical analysis reveals a significant \
    difference between groups A and B (p=0.032 < 0.05), with \
    group A showing a higher mean value of 42.7 compared to group \
    B's 38.1"  
    }}  

    Constraints:  
    - MUST use valid JSON.  
    - DO NOT invent new parameters.  
\end{verbatim}
This prompt format is designed to ensure precise, accountable, and verifiable task execution during the collaboration phase of a multi-agent robotic system. By structuring the prompt around a clearly defined service role and a standardized request context, it enables each agent to focus on task decomposition and execution within its domain of expertise. The explicit instruction to parse task details and flag ambiguities supports robustness against misinterpretation, thereby enhancing reliability. Moreover, mandating structured output in valid JSON ensures machine-readability and seamless integration into automated pipelines. The inclusion of a detailed, explanatory response facilitates transparency and auditability of decisions, which is critical for maintaining trust and traceability in distributed autonomous systems.

At the end of this pipeline, worker are required to complete everything in their request stack, first the collaboration from peers and then the request from leader to finish their assigned subtasks. The process is aligned with the following prompt.
\begin{verbatim}
    Role: AI Planning Expert specialized in HTN planning.  
    Task: Generate a detailed state transition tree that maps \
    possible future states and their outcomes based on available \
    actions, while considering previous HTN.

    Input Context:  
    {{  
    "task_description": "{str_task_desc}",  
    "current_state": {str_cur_state},  
    "available_actions": {str_act_list},  
    "observations": {str_obsv},  
    "htn": {str_htn}  
    }}  

    Instructions:  
    1. **At-Most-Five-Layer State Transition Tree**:  
    - Start from current_state as root node  
    - Expand HTN to predict state transitions in tree structure  
    - Consider transition probabilities for each branch  
    - Factor in current observations  
    2. **State Scoring**:  
    - Evaluate each state's alignment with goal  
    - Score range: 0 (poor) to 1 (optimal)  
    - Consider:  
        - Goal proximity  
        - Transition possibility  
        - Safety constraints  
        - Resource efficiency  
    3. **Action-State Tree Generation**:  
    - Build tree with states as nodes and actions as edges  
    - Evaluate each state for goal conditions  
    - Prune invalid or unsafe branches  
    4. **Output Format**:  
    Return JSON with structure:  
    {{  
        "next_state": {{  
        "state": "<current_state>",  
        "score": <float>,  
        "is_goal": <boolean>,  
        "transitions": [  
            {{  
            "action": "<action_name>",  
            "probability": <float>,  
            "next_state": {{  
                "state": "<resulting_state>",  
                "score": <float>,  
                "is_goal": <boolean>,  
                "transitions": [  
                {{  
                    "action": "<action_name>",  
                    "probability": <float>,  
                    "next_state": {{  
                    "state": "<resulting_state>",  
                    "score": <float>,  
                    "is_goal": <boolean>,  
                    "transitions": []  
                    }}  
                }},  
                ...  
                ]  
            }}  
            }},  
            ...  
        ]  
        }}  
    }}  

    Constraints:  
    - Return ONLY valid JSON  
    - Return empty transitions array for goal states or leaf nodes  
    - All scores MUST be 0-1 range  
    - Consider ONLY valid actions from available_actions  
\end{verbatim}
This prompt is designed to support the final integrative phase of task execution in a multi-agent planning system by leveraging HTN principles to simulate and evaluate potential state trajectories. By instructing agents to construct a bounded-depth state transition tree rooted in the current state and shaped by valid actions, observations, and hierarchical plans, the design promotes foresight and strategic reasoning. Incorporating probabilistic transitions and state scoring ensures that the agent not only considers feasible outcomes but also prioritizes those aligning with task goals, safety, and efficiency. Structuring the output as a recursive JSON tree enforces a standardized, machine-readable format that facilitates downstream interpretation, pruning, and execution. This format allows agents to synthesize inputs from earlier phases—peer collaborations and leader instructions—into coherent action plans while maintaining traceability and adaptability in complex task environments.

For long-horizon tasks and multimodal, complex tasks under dynamic, partial-knowledge settings, we implemented an additional simpler mechanism to predict and verify the next action. The mechanism serves as an individual workflow that does not interfere with the group-multi-agent workflow, thereby preserving an element of independence within the framework. We call it action selector, also an optional third party negotiation agent. The prompt is as follows.
\begin{verbatim}
    Role: Action selection agent  
    Task: Analyze the current context and select the most \
    appropriate action that advances toward the task objective \
    while maintaining safety and efficiency.  

    Input Context:  
    {{  
    "task_description": "{str_task_desc}",  
    "current_state": {str_cur_state},  
    "available_actions": {str_act_list},  
    "observations": {str_obsv}  
    }}  

    Instructions:  
    1. **Action Selection**:  
    - Analyze current state and task objectives  
    - Evaluate available actions for feasibility and safety  
    - Select the single best action for current context  
    - Reason should be concise  
    2. **Output Format**:  
    Return JSON with structure:  
    {{  
        "selected_action": "<action_name>",  
        "reason": "<brief explanation for selection>"  
    }}  

    Constraints:  
    - Return ONLY valid JSON  
    - Selected action MUST be from available_actions list  
    - Reason should be a brief, clear explanation  
\end{verbatim}
This prompt is designed to address the challenges posed by long-horizon and multimodal tasks under dynamic, partially observable environments by isolating a lightweight, context-aware decision-making mechanism. By structuring the agent’s role to operate independently of the broader multi-agent coordination framework, this design preserves modularity and autonomy, allowing for responsive, localized action selection without introducing systemic dependencies. The prompt enforces disciplined reasoning and output through a constrained format that emphasizes safety, feasibility, and alignment with task objectives. Requiring selection from a predefined action set and a concise justification ensures both interpretability and traceability, which are critical for iterative verification and adaptability in complex, evolving task environments.

\section{Open Source Repository}
\rev{The source code and related materials for this work are available at \href{https://anonymous.4open.science/r/HiBerNAC-0E2F/}{\nolinkurl{https://anonymous.4open.science/r/HiBerNAC-0E2F/}}.} This repository is provided to facilitate reproducibility and to support further research in this area.




\printcredits

\bibliographystyle{unsrt}

\bibliography{main_HiBerNAC_v3}





\end{document}